\documentclass[11pt]{article}

\usepackage[margin=1in]{geometry}
\usepackage[numbers,sort&compress]{natbib}
\usepackage{amsmath}
\usepackage{amssymb}
\usepackage{url}
\usepackage{float}
\usepackage{booktabs}
\usepackage{graphicx}            
\usepackage{caption}             
\usepackage{authblk}             
\usepackage[table]{xcolor}
\definecolor{lightgray}{gray}{0.9}

\usepackage[final]{changes}      
\definechangesauthor[name={Reviewer 1}, color=red]{R1}

\providecommand{\keywords}[1]{\vspace{0.5em}\noindent\textbf{Keywords:} #1}
\providecommand{\Vec}[1]{\vec{#1}}

\title{MolMiner: Toward Controllable, 3D-Aware, Fragment-Based\\ Molecular Design}
\author[1,2]{Raul Ortega-Ochoa\thanks{Corresponding author: mail.raulortega@gmail.com}}
\author[1,2]{Tejs Vegge}
\author[3]{Jes Frellsen}
\affil[1]{Department of Energy Conversion and Storage, Technical University of Denmark, DK-2800 Kgs.\ Lyngby, Denmark}
\affil[2]{CAPeX Pioneer Center for Accelerating P2X Materials Discovery}
\affil[3]{Department of Applied Mathematics and Computer Science, Technical University of Denmark}
\date{}

\begin{document}
\maketitle


\begin{abstract}
We introduce MolMiner, a fragment-based, geometry-aware, and order-agnostic autoregressive model for molecular design. MolMiner supports high-dimensional conditional control over twelve physicochemical and structural properties from partial specifications, constructs molecules via symmetry-aware fragment attachments, and conditions each generation step on force-field-relaxed three-dimensional geometry of the partial structure. Conditional control emerges without auxiliary property losses. On targeted property windows, conditioning lifts hit rates by up to 5.25$\times$ over unconditional generation and 3.5$\times$ over the training distribution itself --- overriding the model's intrinsic biases --- at the cost of a small reduction in unconditional distributional fidelity. MolMiner unifies dynamic geometry, symmetry handling, order-agnostic generation, and scalable multi-property conditioning within a single framework.
\end{abstract}

\section{Introduction}
Deep generative models are increasingly central to modern high-throughput screening (HTS) pipelines~\citep{Westermayr2023, htsfunnel2}, where they generate candidate molecules tailored to specific properties before being filtered through successively more expensive stages ranging from machine-learning surrogates~\citep{schutt2017schnetcontinuousfilterconvolutionalneural, gilmer2017neuralmessagepassingquantum, NIPS2015_f9be311e, Kearnes2016-pp} to quantum chemical calculations such as density functional theory~\citep{PhysRev.140.A1133}. These models span a wide range of molecular representations (\emph{e.g.}, strings, graphs, or point clouds) and generative approaches (\emph{e.g.}, VAEs~\citep{chemicalvae, Lim2018}, diffusion~\citep{hoogeboom2022equivariant, NEURIPS2022_eccc6e11}). 
While many methods address isolated challenges ---\emph{i.e.}, chemical validity, structural diversity, or property control--- unconstrained models are often prone to exploring chemically implausible regions of the molecular space, particularly when training signal is limited or representational ambiguity is high. In such conditions, further inductive biases and architectural constraints play a critical role in guiding the model's optimization towards reliable and physically meaningful exploration.

One way to impose such bias is through multi-step, interpretable generation processes operating over molecular fragments rather than atoms, embedding coarse chemical units directly into the generative process. Multi-step fragment-based generation enables flexible control over molecular size and allows chemical validity to be explicitly enforced throughout generation, while supporting human-in-the-loop design. When structure-dependent properties are targeted, incorporating three-dimensional geometry is essential—yet few autoregressive frameworks integrate dynamic geometric information during generation~\citep{voloboev2024reviewfragmentbasednovo2d}.

In addition, many autoregressive models rely on a fixed rollout order, typically growing molecules from a predefined atom or fragment and following a single, predetermined construction trajectory. Such an ordering is not intrinsic to molecular graphs, which admit multiple valid construction sequences. An order-agnostic rollout strategy, in which growth can proceed from any valid fragment while maintaining a single connected graph, better reflects molecular structure and acts as an effective form of data augmentation and regularization, improving generalization.

Finally, conditional generation is critical for use in HTS pipelines, where desired molecular properties are specified upfront. While most models support only single-target conditioning, we enable multi-property control over twelve physicochemical and structural properties. Users can condition on any subset of properties, and the remaining ones are automatically sampled to facilitate efficient, targeted exploration of chemical space.

We introduce MolMiner, a fragment-based, geometry-aware, and order-agnostic generative model designed to impose strong inductive structure on molecular generation. MolMiner operates over chemically meaningful fragments, enforces chemical validity throughout multi-step generation, and incorporates dynamic three-dimensional geometry during rollout. By avoiding a fixed construction order and exposing the model to multiple valid rollouts of the same molecule, we show improved generalization as demonstrated in Figure \ref{fig:train-curve-resample}. Together, these design choices enable reliable, high-dimensional conditional generation. While MolMiner is motivated by high-throughput screening pipelines, this work focuses on evaluating the generative model itself, in terms of validity, diversity, and conditional control.

\section{Related Work}
Our work builds on fragment-based molecular generation approaches such as JTNN~\citep{jin2019junction} and HierVAE~\citep{jin2020hierarchical}, which assemble molecules sequentially while enforcing chemical validity. Like these models, we use coarse-grained molecular fragments and an autoregressive decoding process. Our model is also order-agnostic, similar in spirit to G-SchNet~\citep{Gebauer2022}, allowing flexible rollout without fixing a starting point or strict atom ordering, whereas JTNN and HierVAE are fragment-based but order-fixed, and G-SchNet is order-agnostic but atom-based. Additionally, unlike G-SchNet, we allow the geometry of the partial molecule to remain dynamic during generation, rather than freezing atom positions prematurely. We also explicitly introduce a systematic method to handle fragment symmetries during attachment, an aspect not clearly detailed in earlier fragment-based models such as MoLeR~\citep{maziarz2024learningextendmolecularscaffolds}. Finally, we demonstrate conditional generation across twelve molecular properties simultaneously; a scale of multi-target control that, to the best of our knowledge, has not previously been achieved in molecular generative modeling.

\section{Method}
We model molecular generation as a fragment-based, order-agnostic~\citep{pmlr-v32-uria14, hoogeboom2022autoregressive}, autoregressive process.\\Molecules are decomposed into ring- or bond-based fragments, with attachment points standardized to account for symmetries. Generation proceeds sequentially: at each step, the model is queried with a focal attachment site on the current partial structure and predicts either a fragment to attach or a decision to terminate that site. To incorporate three-dimensional information, the partial molecular structure is relaxed using a force field (\emph{e.g.}, UFF~\citep{uff}), and the resulting geometry is used to inform the prediction. This avoids the rigid, frozen geometries adopted by prior methods~\citep{Gebauer2022} and ensures that predictions are conditioned on physically plausible intermediate structures.

Formally, we define the probability of a molecule $\mathcal{M}$ as the expected likelihood over all valid rollout trajectories $\mathcal{R}(\mathcal{M})$, each consisting of a sequence of fragment attachment actions. At each step, the model predicts a fragment–attachment pair conditioned on the current partial structure and optional conditioning information such as target properties. The generation process also allows a special termination action, which marks an attachment site as closed.

\begin{equation}
p(\mathcal{M}) = \mathbb{E}_{R \sim \mathcal{U}(\mathcal{R}(\mathcal{M}))} \left[ \prod_{i=1}^{|R|} p_\theta\big(x_{i}^{(R)} \big| \mathbf{x}_{<i}^{(R)}, \mathbf{c}\big) \right] ,
\label{eq:order-agnostic}
\end{equation}

Here, $x_i = (f_i, a_i)$ denotes a fragment–attachment pair, where $f_i \in \mathcal{V}_f$ is a fragment from the vocabulary and $a_i \in \mathcal{V}_a(f_i)$ is a valid attachment configuration for $f_i$. The sequence $\mathbf{x}_{<i}^{(R)}$ represents the partial structure up to step $i$, $\mathbf{c}$ denotes optional conditioning information (\emph{e.g.}, target properties), $|R|$ is the rollout length, and $\mathcal{U}$ is the uniform distribution over valid rollout orders.

\begin{figure}[ht]
\centering
\includegraphics[width=0.75\linewidth]{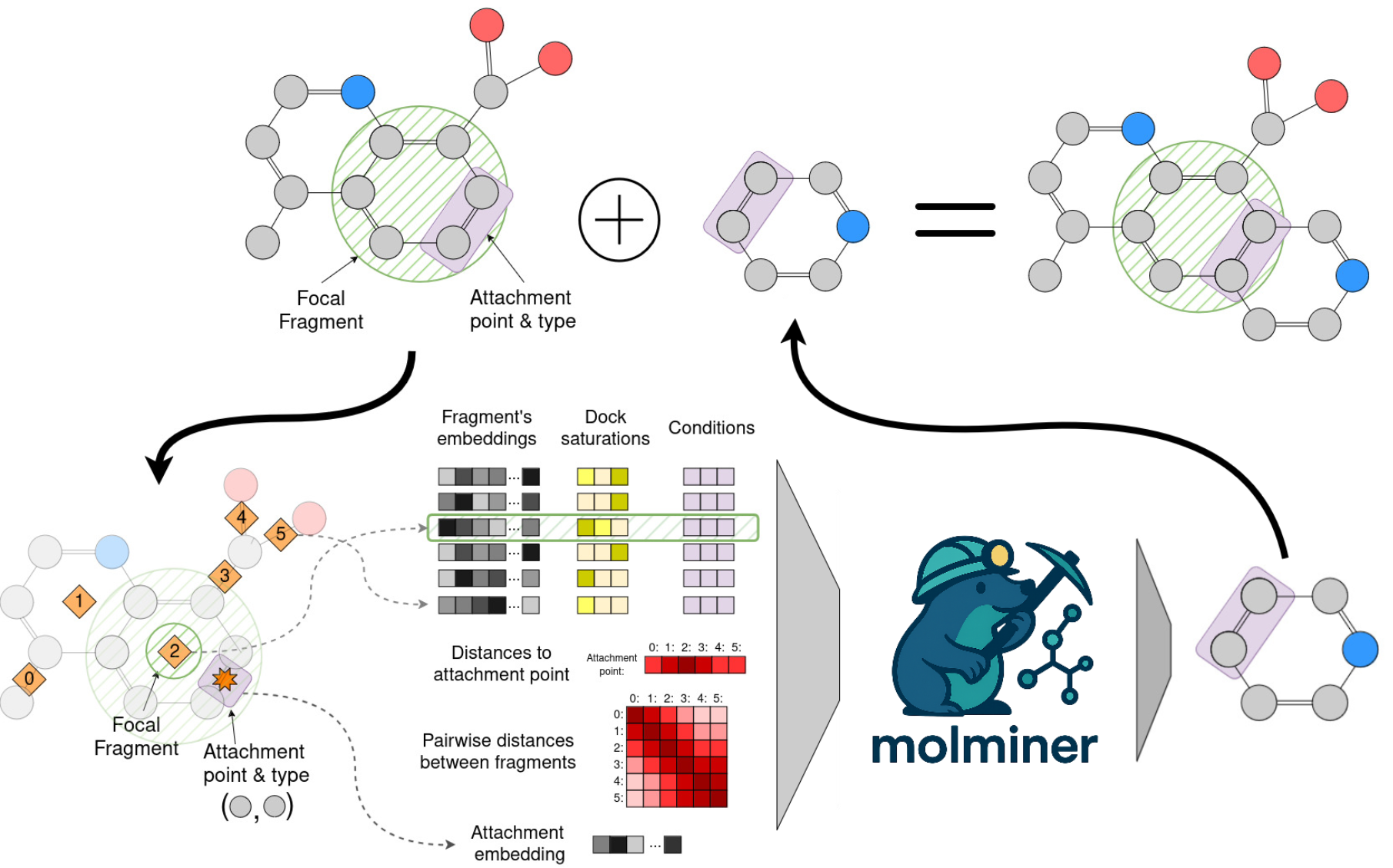}
\caption{Schematic of MolMiner’s fragment-based rollout process. Given a partially grown molecule and a selected focal attachment site, the model predicts the next fragment and attachment configuration in an autoregressive manner. Rollouts proceed in an order-agnostic manner, with growth initiated by an auxiliary predictor that selects the starting fragment.}
\label{fig:molminer-sketch}
\end{figure}

This formulation ensures that the likelihood assigned to a molecule is invariant to the specific sequence of fragment attachments used to construct it. Generation proceeds by alternately attaching fragments or terminating open sites until all sites are resolved, yielding a chemically valid, fully assembled molecule. A schematic illustration of a single step in this autoregressive rollout process is shown in Figure~\ref{fig:molminer-sketch}.

\subsection{Fragment-Based Molecular Representation}
\label{sec:fragmentization}
Molecules naturally exhibit hierarchical structure, often containing repeating substructures such as rings and functional groups. We apply a coarse-graining procedure that decomposes each molecule into a set of fragments corresponding to rings, identified via RDKit’s Smallest Set of Smallest Rings (SSSR)~\citep{rdkit}, and isolated bonds not within a ring. This decomposition strategy is similar to the ``small motif'' variant explored in HierVAE~\citep{jin2020hierarchical}, where molecules are fragmented into minimal cyclic and bond-based motifs.

Each extracted fragment is uniquely represented by its Canonical SMILES string~\citep{smiles,cansmiles}, computed using RDKit’s implementation~\citep{rdkit}, providing a compact, human-readable encoding that is invariant to atom indexing within this scheme\footnote{We note that canonicalization procedures differ between cheminformatics toolkits, and different implementations (e.g., RDKit, OpenBabel~\citep{OBoyle2011}) may produce distinct canonical SMILES for the same molecule~\citep{Dashti2017, doi:10.1021/acs.jcim.5b00543}. Throughout this work, all SMILES are generated using RDKit to ensure consistency and reproducibility.}.
However, canonical SMILES do not retain explicit information about how fragments were connected in the original molecule. To preserve attachment information, we track the mapping between each atom’s original index in the full molecule and its local index in the extracted fragment. This allows us to recover the attachment points necessary for reassembling molecules from fragments and sets the foundation for our symmetry-aware attachment modeling described next.

We treat each fragment as a discrete token, analogous to tokenization in natural language processing (NLP). This abstraction enables us to associate each fragment with a learnable embedding and formulate molecular generation as a stepwise prediction over a sequence of fragment tokens.

\subsection{Symmetry-Aware Attachment Modeling}
Although SMILES syntax allows for explicit encoding of atom-specific metadata, such as attachment points via atom-map numbers, incorporating such information would interfere with the canonicalization procedure itself and alter the resulting SMILES representation. Moreover, explicit labeling does not resolve fragment symmetries, where multiple attachment sites may be chemically indistinguishable. A simple example is benzene, in which all carbon atoms are symmetry-equivalent.

To ensure that fragment attachments are represented consistently and unambiguously, we introduce a symmetry-aware attachment standardization procedure. Because our coarse-graining process extracts fragments corresponding exclusively to rings and isolated bonds—each forming a single cycle—any reindexing induced by canonicalization is restricted to cyclic permutations of atom indices. This structural constraint allows us to explicitly enumerate and resolve all symmetry-equivalent attachment configurations.

In practice, we recover valid index correspondences by comparing local atom environments and identifying cyclic permutations consistent with the fragment’s chemical graph. From these equivalence classes, we construct a canonical reference mapping that assigns attachment points in a deterministic and\\symmetry-invariant manner. This ensures that fragment attachment decisions are invariant to atom indexing and fragment automorphisms, enabling consistent conditioning and generation across symmetric cases. Full algorithmic details and illustrative examples are provided in the Supplementary Information (SI).

\subsection{Order-agnostic Molecular Rollouts}
\label{sec:order-agnostic}
Molecular generation is formulated as a sequence of fragment attachment decisions conditioned on a focal attachment site. Unlike prior approaches that impose a fixed traversal order (\emph{e.g.}, breadth-first or depth-first), we adopt an order-agnostic rollout strategy: at each step, the focal attachment site is sampled uniformly from the set of currently open sites. The only constraint is that new fragments must attach to the existing structure, ensuring that the molecule grows as a single connected component. This design avoids biases induced by a predefined ordering and allows multiple valid rollout trajectories for the same molecule.

Rollouts are initialized by selecting a starting fragment and inserting its available attachment points into a queue. At each step, a focal site is selected from this queue and the model predicts either a fragment to attach or a termination decision for that site. Termination is local rather than global: attachment points are closed independently, and generation concludes only when all open sites have been resolved. This decentralized termination mechanism reflects the graph-structured nature of molecules and enables flexible, non-linear growth.

During training, rollout sequences and intermediate geometries are precomputed, enabling efficient learning without repeated force-field optimization. During generation, in contrast, molecules are constructed incrementally, with the geometry relaxed after each attachment step using a classical force field. This ensures that autoregressive predictions remain conditioned on physically plausible intermediate structures throughout sampling.

\subsection{Model Architecture}
We implement the model as a decoder-only transformer~\citep{vaswani2023attentionneed} operating over a set of fragment tokens. Each fragment is represented by a learnable embedding, which is augmented by concatenation with three normalized scalar features encoding the fraction of attachment sites that are occupied, free, or sealed. These features provide local chemical context and allow the model to distinguish between fully bonded, partially open, and terminated fragments.

To make the model geometry-aware, we incorporate spatial information directly into the attention mechanism via a global attention bias~\citep{shehzad2024graphtransformerssurvey}. Rather than relying on fixed positional encodings, attention between fragments is modulated by their three-dimensional proximity. Concretely, fragment centroids are used to compute a distance-based kernel that biases attention scores, with a learnable scalar controlling the strength of the geometric contribution. We note that MolMiner does not explicitly generate 3D coordinates, but instead incorporates geometry as an attention bias derived from force-field-relaxed intermediate structures. Because molecular structures lack a canonical linear order, this geometry-based bias provides a structure-aware alternative to positional embeddings while remaining invariant to fragment ordering. Specifically, the attention coefficients between fragments $i$ and $j$ are defined as
\begin{equation}
\alpha_{ij} = \frac{e^{ g(h_i, h_j) + \theta \cdot D_{ij}}}{\sum_{k=1}^{N} e^{ g(h_i, h_k) + \theta \cdot D_{ik}}}, \quad
D_{ij} = e^{-\frac{\|\mathbf{x}_i - \mathbf{x}_j\|^2}{2\sigma^2}}, \quad
g(h_i, h_j) = \frac{h_i h_j^\top}{\sqrt{d_h}} ,
\end{equation}
where $\mathbf{x}_i$ denotes the centroid of fragment $i$, $h_i$ is the hidden representation of fragment $i$ produced by the previous transformer layer, $D_{ij}$ is a Gaussian-decayed distance kernel, and $\theta$ is a learnable scalar controlling the strength of the geometric bias. Because the attention bias depends only on pairwise distances between fragment centroids, the resulting representation is invariant to global translations and rotations of the molecular geometry. This mechanism allows the model to incorporate three-dimensional structural information without relying on explicit positional encodings.

During generation, the transformer is conditioned on the current set of fragments, a designated focal fragment, and a specific attachment site (the \emph{hit location}). After processing the fragment set, we perform a focalized readout in which the focal fragment aggregates information from all other fragments, with attention further biased toward fragments closer to the hit location. This produces a context vector that emphasizes the local growth site while retaining global structural information. The resulting representation is concatenated with the conditioning properties and projected onto the vocabulary of fragment–attachment actions, including the termination action.

\subsection{Training Objective}
We train the model to maximize the log-likelihood of each molecule $\mathcal{M}$ under the order-agnostic rollout factorization~\citep{pmlr-v32-uria14, hoogeboom2022autoregressive}, conditioned on target properties $\mathbf{c}$:
\begin{equation}
\mathcal{L}(\theta \mid \mathcal{M}) = 
\log \mathbb{E}_{R \sim \mathcal{U}(\mathcal{R}(\mathcal{M}))} \left[ \prod_{i=1}^{|R|} p_\theta\big(x_{i}^{(R)} \big| \mathbf{x}_{<i}^{(R)},\mathbf{c}\big) \right] \geq
\mathbb{E}_{R \sim \mathcal{U}(\mathcal{R}(\mathcal{M}))} \left[ \sum_{i=1}^{|R|} \log p_\theta\big(x_{i}^{(R)} \big| \mathbf{x}_{<i}^{(R)},\mathbf{c}\big) \right]
\end{equation}
The expectation is over all valid rollouts $R$ of $\mathcal{M}$, with the lower bound derived via Jensen’s inequality~\citep{jensen1905}. In practice, we use a Monte Carlo approximation of the expectation and randomly sample one rollout per molecule per epoch, providing natural data augmentation by exposing the model to diverse construction orders. At each step, the model predicts the next fragment-attachment pair or a termination action, conditioned on the current partial structure and target properties.

To initiate rollouts, we jointly train an auxiliary model to predict a suitable starting fragment from the target properties. This predictor is a feed-forward network that outputs independent probabilities for each fragment in the vocabulary, framing the task as multi-label classification. It is trained with binary cross-entropy loss to encourage high scores for fragments present in the molecule. Both models share the same training splits.

Together, these components enable end-to-end conditional generation—from fragment selection to flexible, geometry-aware rollouts. Importantly, conditioning is implemented in a fully implicit manner: target properties are provided as inputs during training, but no auxiliary loss is applied to enforce property compliance. This allows the model to learn property alignment organically from the data distribution.

\subsection{Sampling Procedure}
\label{sec:sampling-explanation}
To generate molecules conditioned on user-specified properties, we first complete the conditioning vector when only a subset of target properties is provided. Missing properties are sampled from the empirical joint distribution of training properties using a Gaussian Mixture Model (GMM)~\citep{mclachlan2000finite}. Conditioning values are drawn from the corresponding conditional distribution given the observed properties, ensuring that completed conditioning vectors remain realistic and consistent with the training data. Full details of the GMM formulation and conditional sampling procedure are provided in the SI.

Once the conditioning vector is completed, generation is initialized by selecting a starting fragment. A trained fragment predictor assigns independent probabilities over the fragment vocabulary conditioned on the target properties, from which a seed fragment is sampled. The molecule is then constructed autoregressively via the order-agnostic rollout procedure described above.

We evaluate several sampling strategies in the SI, including greedy and probabilistic decoding, as well as restricting seed fragment selection to the top-$k$ predictions (with $k=3,5,10$). We further analyze how conditioning values influence the choice of starting fragment.

\section{Experiments}
Our evaluation focuses on intrinsic generative performance rather than downstream task optimization. In particular, we assess distributional fidelity and controllability, which are key requirements for integration into larger discovery pipelines.
We evaluate our model on a subset of the ZINC dataset~\citep{doi:10.1021/ci3001277} originally curated for ChemicalVAE~\citep{chemicalvae}, containing approximately 200,000 drug-like molecules. Each molecule is annotated with 12 physicochemical properties computed using RDKit, which are used both for conditioning and evaluation. These properties include: logP (logarithm of water partition coefficient, used as a measure of lipophilicity)~\citep{Wildman1999PredictionOP}, QED (quantitative estimate of drug-likeness)~\citep{Bickerton2012}, SAS (synthetic accessibility)~\citep{Ertl2009}, FractionCSP3 ($sp^{3}$ carbon fraction), molecular weight, TPSA (topological polar surface area)~\citep{doi:10.1021/jm000942e}, MR (molar refractivity, descriptor accounting for molecular size and polarizability)~\citep{Wildman1999PredictionOP}, hydrogen bond donors and acceptors (HBD, HBA), ring count, number of rotatable bonds (flexibility), and number of chiral centers (stereochemical complexity), (see SI for details). We adopt an 80/10/10 train/validation/test split.

\begin{figure}[h]
    \centering
    \includegraphics[width=0.8\linewidth]{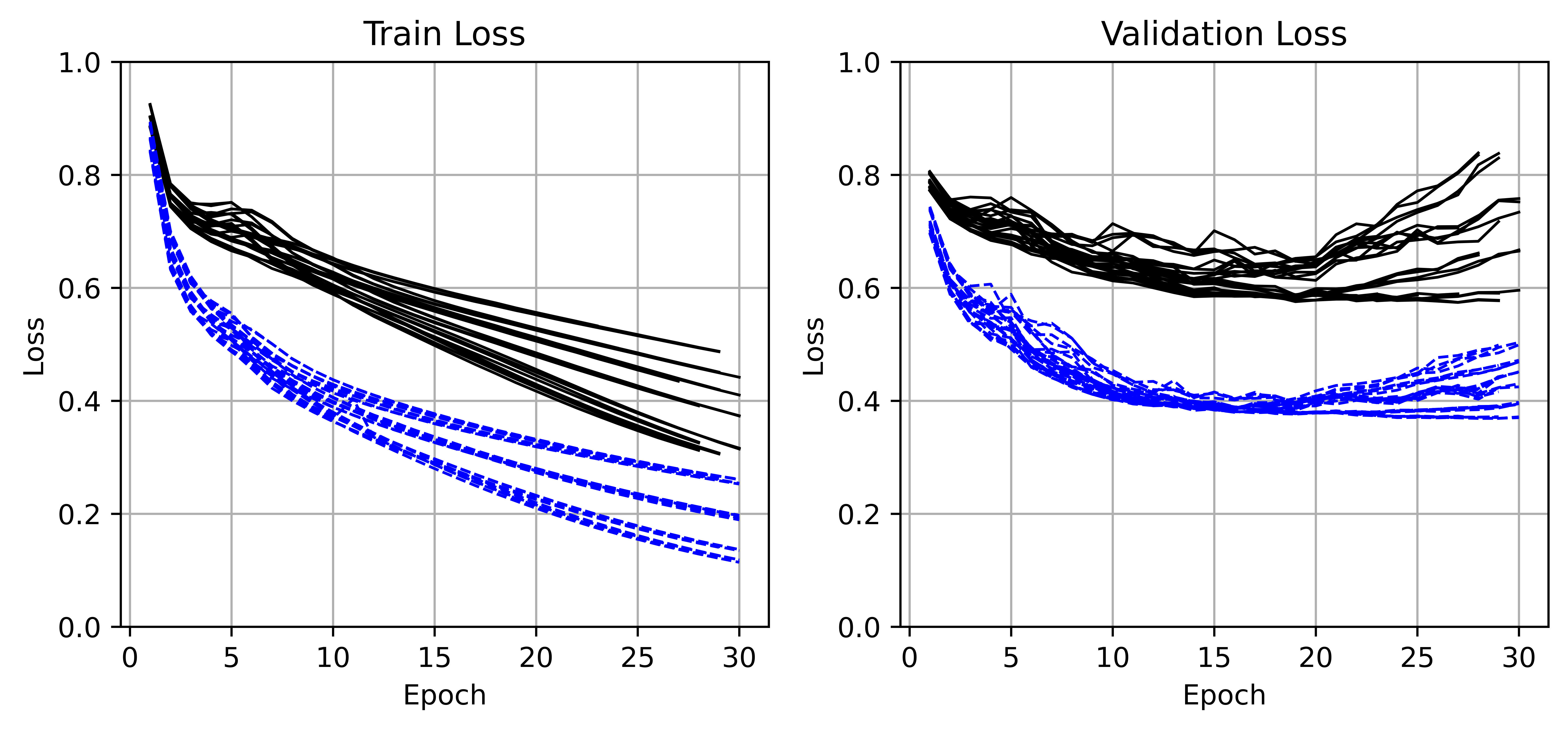} \caption{Training and validation curves for models trained with 3 (black) and 12 (blue) conditioning properties. Models with 12 conditions consistently achieve lower training and validation loss, demonstrating improved generalization.}
    \label{fig:train-curve-tomographic}
\end{figure}

\paragraph{Effect of Conditioning Dimensionality.}
We first investigate how the number of conditioning properties affects generative performance. To this end, we train otherwise identical models conditioned on either 3 or 12 molecular properties and track reconstruction loss on training and validation sets. As shown in Figure~\ref{fig:train-curve-tomographic}, models conditioned on 12 properties consistently achieve lower losses throughout training and generalize better than their 3-property counterparts. This gap persists across epochs, indicating that richer non-redundant conditioning provides a more informative constraint on molecular structure~\citep{ortegaochoa2025tomographicinterpretationstructurepropertyrelations}.

\begin{figure}[h]
    \centering 
    \includegraphics[width=0.8\linewidth]{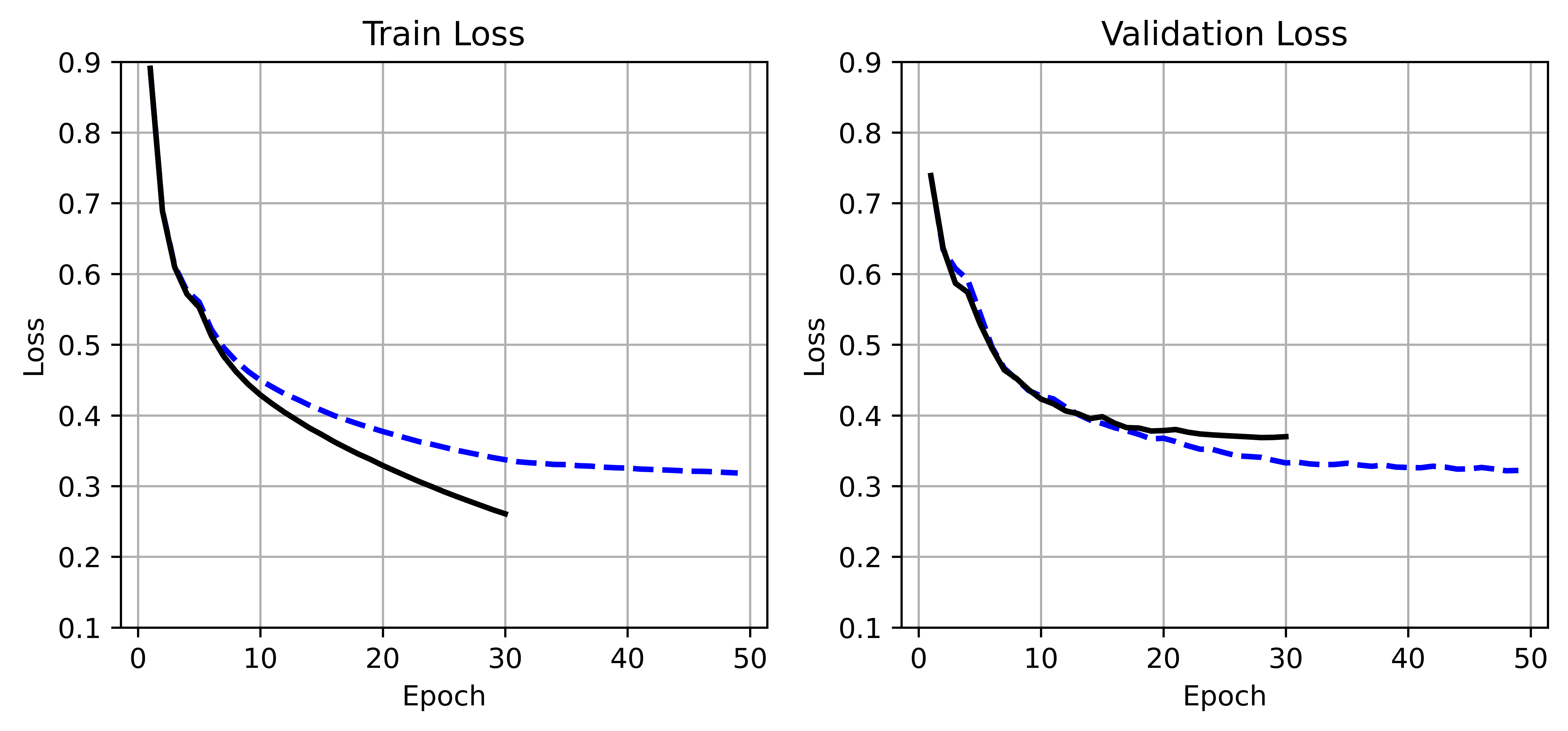} \caption{Training and validation curves comparing a baseline model (black) to a variant trained with rollout resampling (blue). Resampling reduces the train–validation gap and enables continued improvement beyond 30 epochs, indicating improved generalization and regularization.}
    \label{fig:train-curve-resample}
\end{figure}

\paragraph{Effect of Rollout Resampling.}
We next assess the impact of resampling molecular rollouts during training. Instead of fixing a single construction order per molecule, we resample a valid rollout at each epoch, exposing the model to multiple equivalent generation trajectories. As shown in Figure~\ref{fig:train-curve-resample}, rollout resampling leads to lower validation loss and a markedly reduced train–validation gap compared to a baseline trained without resampling. Notably, the resampled model continues improving beyond 30 epochs, whereas the baseline quickly plateaus. These results indicate that rollout resampling acts as an effective form of data augmentation and regularization, improving generalization in the order-agnostic setting.

\paragraph{Final Training Configuration.}
Based on the ablation studies, we adopt full 12-property conditioning, geometry-aware attention with a positively initialized bias, and rollout resampling for all subsequent experiments. The final model is a 4-layer decoder-only transformer trained with AdamW~\citep{kingma2017adammethodstochasticoptimization, loshchilov2019decoupledweightdecayregularization} and a linear warmup–decay schedule. Hyperparameters were selected via grid search (see SI for further details), with the final configuration using a dropout rate of 0.3, 16 attention heads, a warmup ratio of 0.15, and a peak learning rate of $5\times10^{-5}$. Models are trained for approximately 50 epochs.

\subsection{Benchmarking Unconditional Generation}\label{sec:uncond-gen}
We first evaluate MolMiner in an unconditional setting to establish a baseline for distributional fidelity. Performance is assessed by comparing the distributions of twelve molecular properties between approximately 5{,}000 generated molecules and the reference dataset using the 1D Wasserstein distance~\citep{DBLP:journals/corr/abs-1811-12823}. Although computed over one-dimensional marginals, all properties are evaluated on the same generated molecule set; matching multiple correlated marginals therefore provides a practical proxy for joint distributional fidelity while avoiding the statistical instability of high-dimensional Wasserstein estimates.

Because MolMiner is inherently conditional, unconditional generation is simulated by sampling conditioning vectors to match the training distribution from the Gaussian mixture model (GMM). We benchmark against HierVAE~\citep{jin2020hierarchical}, the most comparable fragment-based model in terms of generation strategy and architectural design, GDSS~\cite{pmlr-v162-jo22a} as a representative diffusion-based approach, and exclude MARS~\citep{xie2021marsmarkovmolecularsampling} as it relies on oracle property evaluations during MCMC-based generation and is therefore not directly comparable in practical property-controlled generation settings.

\begin{figure}[ht]
    \centering
    \includegraphics[width=\linewidth]{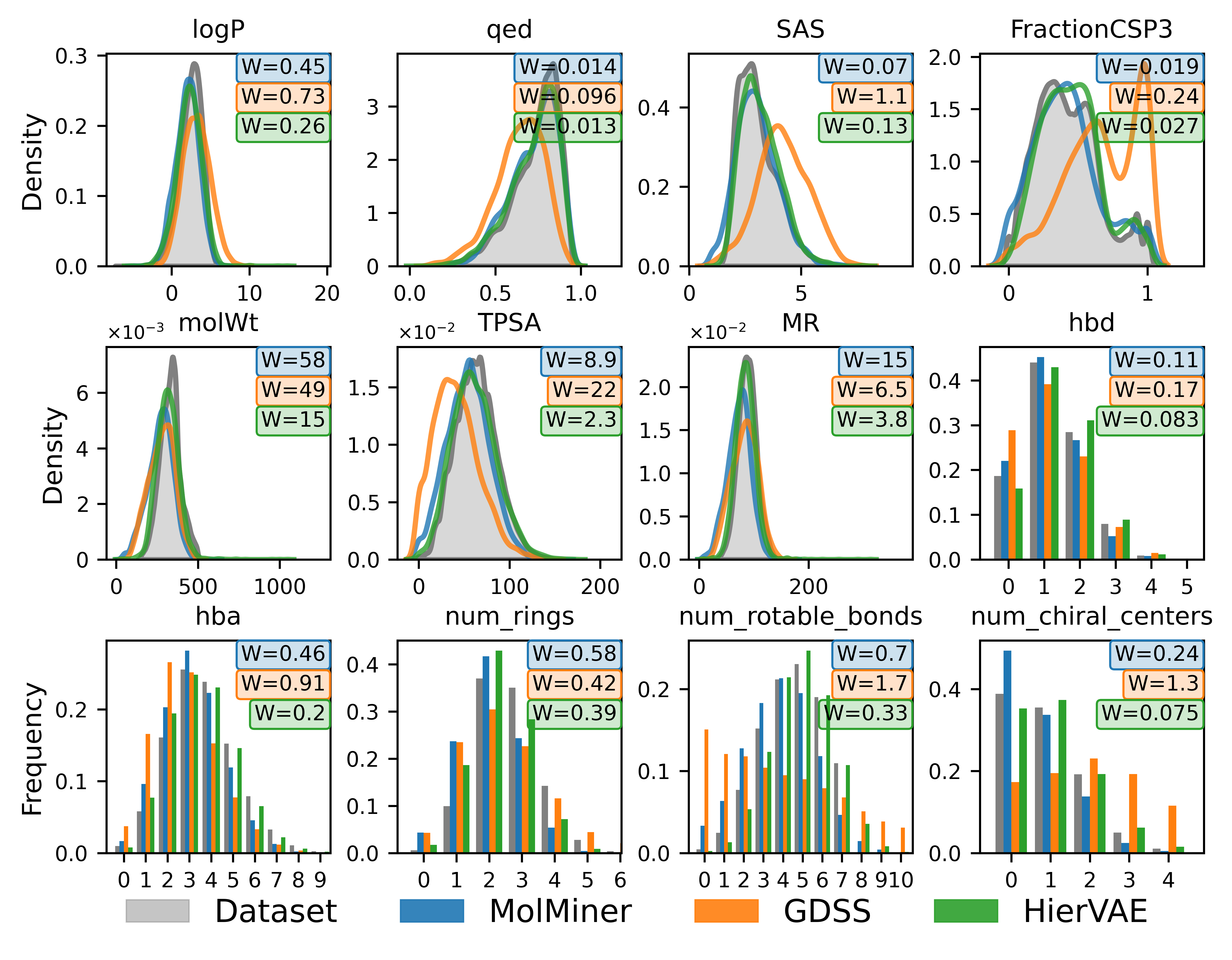}
    \caption{Kernel density estimates (KDE) and histograms of twelve molecular properties across 5{,}000 generated molecules. Distributions from the reference dataset are compared against MolMiner, HierVAE and GDSS.}
    \label{fig:generation-unconditional-benchmark-dists}
\end{figure}

Figure~\ref{fig:generation-unconditional-benchmark-dists} shows kernel density estimates and histograms for all twelve properties, alongside their respective Wasserstein distances. MolMiner closely matches the reference distributions across most dimensions. Differences are most pronounced for molecular weight, molar refractivity (MR), and topological polar surface area (TPSA), which exhibit a slight shift toward smaller values. This effect arises from the order-agnostic rollout formulation: molecule construction involves explicit fragment attachment and termination actions, and because each open attachment site must be closed, termination actions are more frequent than fragment additions during training. This induces a mild bias toward early termination and, consequently, slightly smaller molecules on average. We verified this mechanism through a targeted ablation study, showing that reweighting the loss contribution of the termination token by a factor of 0.5 reduces the small-molecule bias and improves alignment for molecular weight, MR, and TPSA. The results reported here already incorporate this partial correction, confirming that the observed deviations reflect a controllable modeling artifact rather than a fundamental limitation.
\begin{table}[ht]
\centering
\caption{Unconditional generation benchmark across models. Validity is reported under RDKit sanitization checks using the RDKit version in each model's original implementation; for GDSS, validity is computed after largest-component extraction. Frag.\ (\%) denotes the percentage of generated molecules that were fragmented and required extraction of the largest connected fragment prior to evaluation. All metrics computed on $N\approx5{,}000$ generated molecules.}
\label{tab:unconditional}
\resizebox{\textwidth}{!}{%
\begin{tabular}{lccccccc}
\toprule
& \textbf{Validity (\%)}
& \textbf{Frag. (\%)}
& \textbf{Uniqueness (\%)}
& \textbf{Novelty (\%)}
& \textbf{Diversity} \\
\midrule
MolMiner (ours) & 100.0  & 0.00  & 98.5   & 99.9          & 0.8919 \\
HierVAE         & 100.0  & 0.00  & 100.0 & 99.9          & 0.8759 \\
GDSS            & 100.0  & 38.96 & 99.6   & 100.0 & 0.9026 \\
\bottomrule
\end{tabular}}
\smallskip
\begin{minipage}{\textwidth}
\end{minipage}
\end{table}

We additionally report uniqueness, novelty, and diversity (mean pairwise Tanimoto distance). Molecular identity is defined using the first block of the InChIKey~\citep{Heller2015,Pletnev2012}, which captures molecular connectivity while ignoring stereochemistry. As shown in Table~\ref{tab:unconditional}, MolMiner achieves high uniqueness (98.5\%) and novelty (99.9\%), comparable to HierVAE (100.0\% and 99.9\%) and GDSS (99.6\% and 100.0\%), while maintaining perfect validity by construction. MolMiner also exhibits higher diversity than HierVAE (0.8919 vs.\ 0.8759), and remains competitive with GDSS (0.9026). Notably, GDSS produces fragmented outputs in a substantial fraction of cases (38.96\%), requiring extraction of the largest connected component prior to evaluation, a common issue in diffusion-based molecular generative models~\cite{hoogeboom2022equivariant}. In contrast, MolMiner enforces chemical validity throughout generation and produces fully connected molecules by construction.

To complement these external comparisons, we also report an internal specialist-vs-generalist ablation in SI Section~\ref{sec:si-generalistvsspecialist}, in which MolMiner is retrained with all conditioning channels except logP masked. The resulting 1D specialist matches the reference marginals more closely on most properties (e.g., molWt W: 58 $\rightarrow$ 38, TPSA W: 8.9 $\rightarrow$ 3.3), indicating that part of the residual deviation observed for molWt, MR, and TPSA above reflects not only the termination-bias artifact but also the cost of high-dimensional generalist conditioning: a model required to satisfy twelve simultaneous property constraints has less freedom to reproduce the empirical marginals than one conditioned on a single property. MolMiner thus trades a small amount of unconditional distributional fidelity for the ability to control eleven additional property axes simultaneously, a trade we exploit in the targeted multi-property generation experiments of Section~\ref{sec:targeted-newexp}.

Overall, MolMiner demonstrates competitive unconditional generation performance, closely matching reference property distributions and standard quality metrics while maintaining perfect validity. The remaining deviations are small, interpretable, and tied to identifiable aspects of the rollout and conditioning mechanisms. Having established this unconditional baseline, we now turn to the conditional generation setting, where MolMiner’s design choices enable flexible and high-dimensional property control.

\subsection{Benchmarking Conditional Generation}\label{subsec:cond}
To evaluate conditional generation, we perform a systematic, property-by-property stress test that probes how accurately MolMiner responds to target property specifications across a wide dynamic range. For each of the twelve physicochemical and structural properties, target values are uniformly sampled across the range $\mu \pm 2\sigma$ of the empirical training distribution. For each target value, the remaining eleven properties are sampled conditionally from the GMM prior, yielding a realistic twelve-dimensional conditioning vector. To quantify conditional control, we report Mean Absolute Error (MAE) and coefficient of determination ($R^{2}$) between prompted and generated property values.

\begin{figure}[ht]
    \centering
    \includegraphics[width=1.0\linewidth]{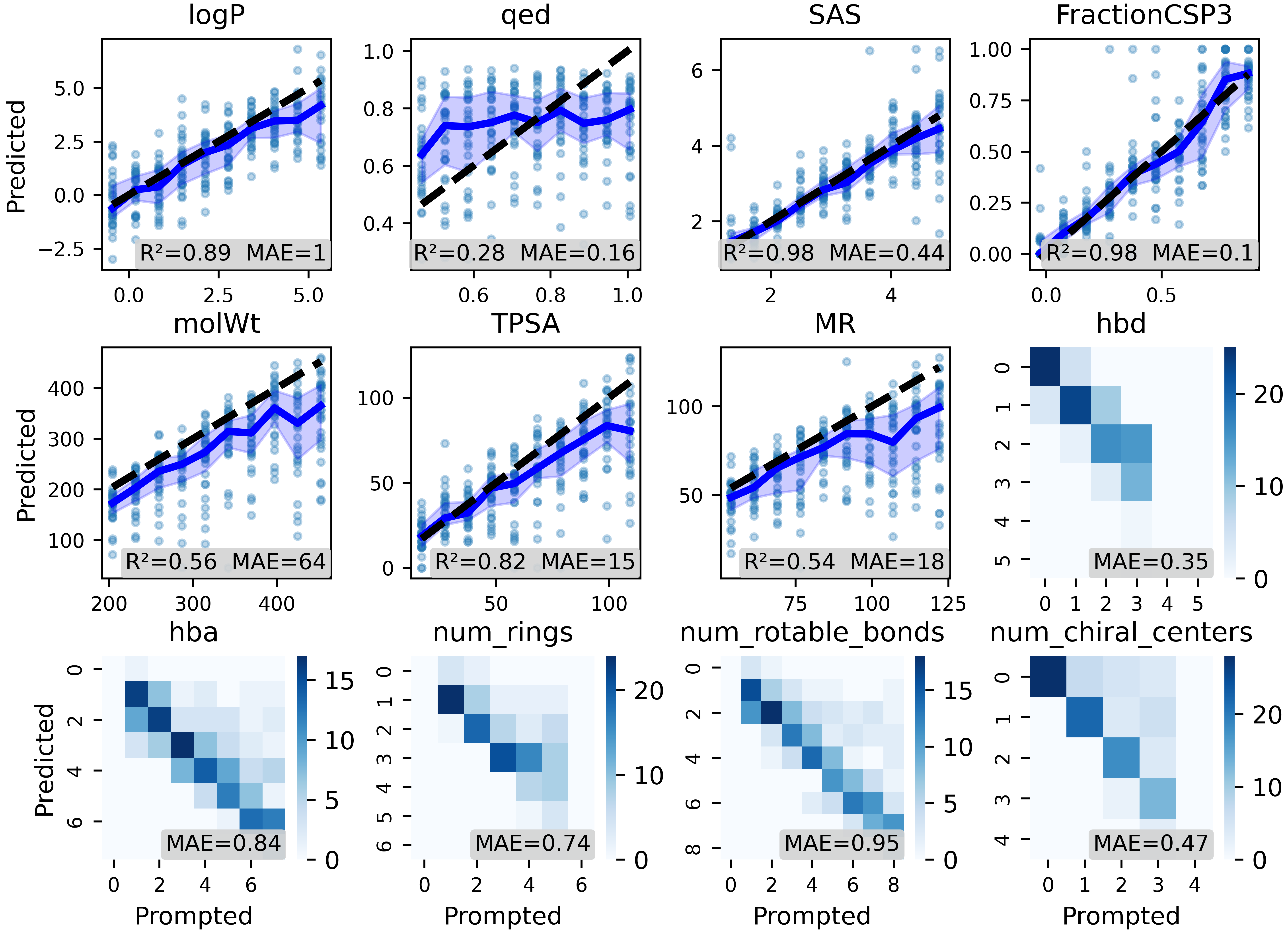}
    \caption{Calibration of conditional generation across twelve molecular properties. Continuous properties show predicted versus prompted values with median and 25--75 quantile bands; discrete properties are summarized as confusion matrices. Mean Absolute Error (MAE) is reported for all properties; for continuous properties, the coefficient of determination $R^2$ is additionally reported between the median predicted and prompted values.}
    \label{fig:generation-calibration}
\end{figure}

Molecules are then generated autoregressively using this completed conditioning vector. This procedure is repeated 30 times per target value, and the reported results reflect aggregate statistics over these realizations. The full process is carried out independently for each property, providing a comprehensive assessment of controllability across all conditioning dimensions.

Calibration plots compare the prompted (target) values with the properties computed from the generated molecules. For continuous properties, we report median predicted values with 25-75 quantile bands, while discrete properties are summarized using confusion matrices.

As shown in Figure~\ref{fig:generation-calibration}, MolMiner achieves low MAE and high $R^2$ across most properties, indicating strong agreement between prompted and generated values. Notably, no auxiliary loss explicitly enforces property values during training. The observed calibration across the twelve properties emerges organically (without explicit loss supervision) from the learned structure–property relationships. QED behaves differently: while the model reproduces the correct marginal distribution for QED in the unconditional setting, conditional control over this composite descriptor is weaker. MolMiner’s behavior with respect to QED remains closer to unconditional generation, albeit with partial responsiveness to conditioning. We hypothesize this reflects QED's nature as a composite descriptor: because QED is largely determined by the eleven other conditioning properties already in the prompt, the marginal information added by the QED channel is small, and the model effectively conditions on QED indirectly through its correlates.

Molecular weight, topological polar surface area (TPSA), and molar refractivity (MR) exhibit small, systematic deviations under conditioning, mirroring the termination-bias effect documented in Section~\ref{sec:uncond-gen}.

To quantify the per-property cost of generalist conditioning, we additionally compare MolMiner against an internal specialist baseline (MolMiner-logP) trained with all conditioning channels except logP masked (SI Section~\ref{sec:si-generalistvsspecialist}). On its single trained property, the specialist improves over the generalist ($R^2 = 0.98$ vs. $0.89$, MAE $= 0.54$ vs. $1.0$ on logP), confirming the expected specialist advantage. However, controllability collapses on the eleven remaining properties: SAS drops from $R^2 = 0.98$ to $0.33$, FractionCSP3 from $0.98$ to $0.47$, TPSA from $0.82$ to near zero, and the discrete properties lose their diagonally-dominant calibration structure. The 12D generalist accepts a small per-property cost in exchange for simultaneous, calibrated control across twelve correlated property axes from partial user specifications.

\begin{figure}[H]
    \centering
    \includegraphics[width=0.9\linewidth]{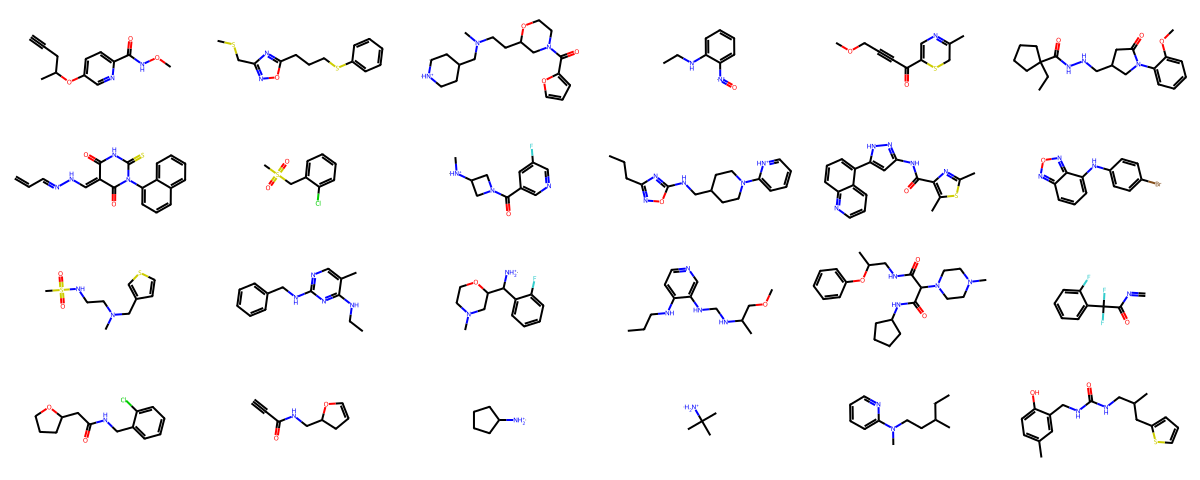}
    \caption{Random sample of 24 molecules generated by MolMiner under unconditional sampling (conditioning vectors drawn from the GMM prior).}
    \label{fig:sup-generation-unconditional-examples}
\end{figure}

\begin{figure}[H]
    \centering
    \includegraphics[width=0.8\linewidth]{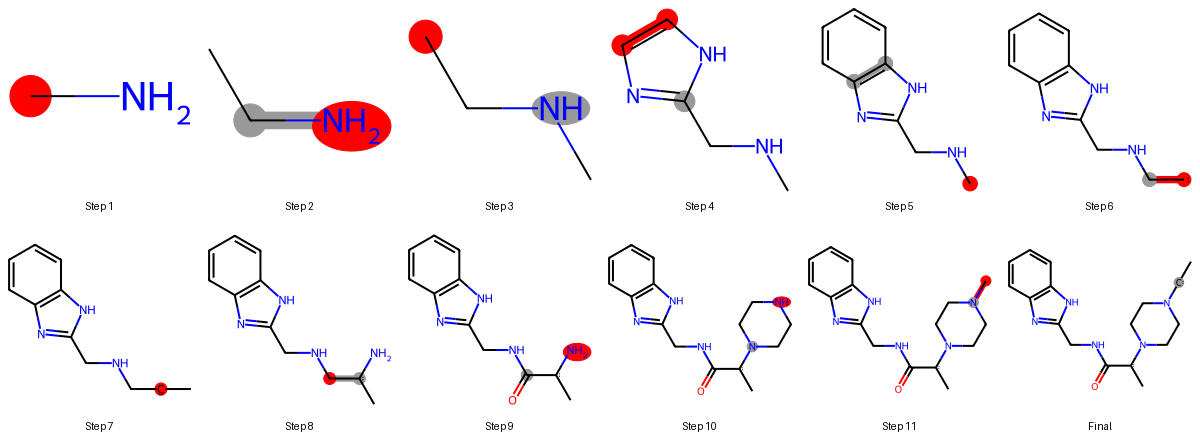}
    \caption{A representative autoregressive generation trajectory from MolMiner, sampled under unconditional generation. Atoms in red indicate the attachment site at which the molecule is grown at each step. Termination (site-closing) steps are omitted for compactness, as they do not change the structure visually.}
    \label{fig:decoding_example}
\end{figure}

Representative examples of generated molecules and a single autoregressive generation trajectory are shown in Figures~\ref{fig:sup-generation-unconditional-examples} and~\ref{fig:decoding_example}, illustrating fragment-based growth.

\subsection{Targeted Multi-Property Generation}\label{sec:targeted-newexp}
While the calibration analysis in Section~\ref{subsec:cond} evaluates conditional control along individual property axes, practical use cases in high-throughput screening typically require generation within 
\emph{multi-property regions} defined by simultaneous constraints. We therefore evaluate MolMiner in a complementary setting: rather than sweeping one property at a time, we define eight canonical property windows spanning standard drug-discovery regions of chemical space and measure the fraction of generated molecules that fall within each. The filters are defined in Table~\ref{tab:filter_defs} and reflect commonly cited cutoffs in the medicinal-chemistry literature; exact thresholds in these regions are not standardized, so we use representative values 
drawn from canonical references.

\begin{table}[h]
\centering
\caption{Property windows used in the targeted-generation experiment. 
A molecule passes a filter if and only if it satisfies all constraints 
simultaneously. Each filter relates to a canonical region of chemical 
space; the cutoffs follow the cited references.}
\label{tab:filter_defs}
\small
\setlength{\tabcolsep}{3pt}
\renewcommand{\arraystretch}{1.1}
\begin{tabular}{lp{0.55\linewidth}l}
\toprule
\textbf{Filter} & \textbf{Constraints} & \textbf{Reference} \\
\midrule
F1 (drug-like)   & MW $\leq$ 500, logP $\leq$ 5, HBD $\leq$ 5, HBA $\leq$ 10 & Lipinski's Rule of 5~\cite{LIPINSKI20013} \\
F2 (lead-like)   & 250 $\leq$ MW $\leq$ 350, 1 $\leq$ logP $\leq$ 3, HBD $\leq$ 3, HBA $\leq$ 6, RotB $\leq$ 7 & Teague \emph{et al.}~\cite{hann_oprea} \\
F3 (fragment)    & MW $\leq$ 300, logP $\leq$ 3, HBD $\leq$ 3, HBA $\leq$ 3, RotB $\leq$ 3 & Inspired by Congreve \emph{et al.}~\cite{CONGREVE2003876} \\
F4 (CNS)         & MW $\leq$ 360, logP $\leq$ 3, TPSA $\leq$ 90, HBD $\leq$ 3 & Inspired by Wager \emph{et al.}~\cite{wager} \\
F5 (3D-rich-rigid) & FractionCSP3 $\geq$ 0.5, RotB $\leq$ 4, num\_rings $\geq$ 2 & three-dimensional, rigid scaffolds \\
F6 (easy-synth)  & SAS $\leq$ 2.5 & Ease of synthesis (SAS)~\cite{Ertl2009} \\
F7 (high-QED)    & QED $\geq$ 0.7 & high drug-likeliness \cite{Bickerton2012} \\
F8 (achiral)     & num\_chiral\_centers = 0 & Stereochemical descriptor \\
\bottomrule
\end{tabular}
\end{table}

For each filter, we draw conditioning vectors from the GMM and reject those that do not satisfy the filter criteria, retaining $N \approx 5{,}000$ accepted prompts. One molecule is then generated per accepted prompt with MolMiner, using the same trained model across all windows without retraining, oracle-based filtering, or auxiliary compliance losses. Hit rates are compared against three unconditional baselines (HierVAE, GDSS, and MolMiner without conditioning) and against the training set's natural hit rate as a baseline reference. This procedure tests whether MolMiner's multi-property conditioning, deployed on-the-fly at inference time, translates into operationally relevant control: the ability to steer generation into specific, simultaneously-constrained regions of chemical space without task-specific retraining.

\begin{figure}[!htbp]
\centering
\begin{minipage}[t]{0.60\textwidth}
\vspace{0pt}
\centering
\captionof{table}{Targeted-generation hit rates across property filters. For each filter (rows), the percentage of generated molecules satisfying the filter criteria is reported for: the ZINC training subset (reference, italicized as baseline frequency), three unconditional baselines (HierVAE, GDSS, MolMiner-u, denoting the final MolMiner checkpoint evaluated under unconditional sampling), and MolMiner conditioned on prompts drawn from the filter's target region by Gaussian mixture model rejection sampling (one molecule per prompt). The final column reports the fold-enrichment of conditional over unconditional MolMiner. Per-row maximum in bold. Filter definitions in Table~\ref{tab:filter_defs}. All hit rates computed on $N \approx 5{,}000$.}
\label{tab:targeted_generation}
\resizebox{\linewidth}{!}{%
\setlength{\tabcolsep}{3pt}
\begin{tabular}{lcccccc}
\toprule
& \textbf{\textit{Train}}
& \textbf{HierVAE}
& \textbf{GDSS}
& \textbf{MolMiner-u}
& \textbf{MolMiner}
& \textbf{Fold} \\
\midrule
F1 (drug-like) & \textit{98.5} & 96.7 & 84.5 & 98.3 & \textbf{99.0} & 1.01$\times$ \\
F2 (lead-like) & \textit{26.6} & 26.7 & 12.1 & 25.8 & \textbf{40.6} & 1.57$\times$ \\
F3 (fragment) & \textit{7.1}  & 6.9  & 18.1 & 23.6 & \textbf{81.2} & 3.44$\times$ \\
F4 (CNS) & \textit{43.7} & 50.8 & 41.7 & 65.3 & \textbf{88.1} & 1.35$\times$ \\
F5 (3D-rich-rigid) & \textit{16.2} & 12.8 & 31.2 & 10.9 & \textbf{57.5} & 5.25$\times$ \\
F6 (easy-synth) & \textit{29.4} & 23.2 & 6.9  & 30.1 & \textbf{78.3} & 2.60$\times$ \\
F7 (high-QED) & \textit{67.1} & 63.0 & 37.7 & 61.3 & \textbf{68.7} & 1.12$\times$ \\
F8 (achiral) & \textit{38.8} & 35.3 & 17.3 & 49.4 & \textbf{94.9} & 1.92$\times$ \\
\bottomrule
\end{tabular}}
\end{minipage}%
\hfill
\begin{minipage}[t]{0.38\textwidth}
\vspace{0pt}
\centering
\includegraphics[width=\linewidth]{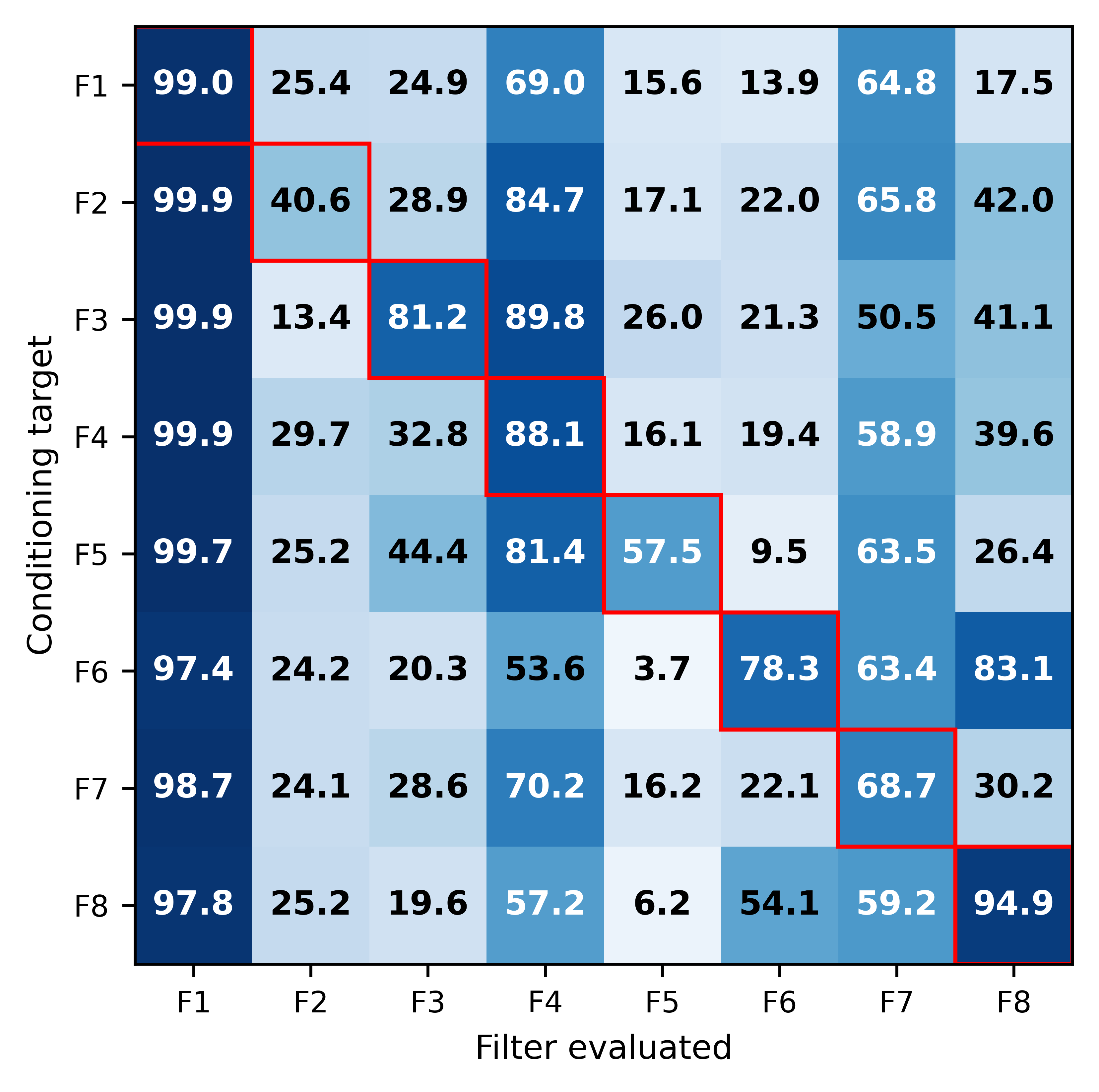}
\captionof{figure}{Selectivity of conditional MolMiner. Rows index the conditioning target; columns the filter evaluated. Diagonal cells (red) correspond to the targeted hit rates in Table~\ref{tab:targeted_generation}.}
\label{fig:selectivity_heatmap}
\end{minipage}
\end{figure}

Table~\ref{tab:targeted_generation} shows that MolMiner exceeds every unconditional baseline on every filter when prompted with a target region drawn from that filter. Fold-enrichment over unconditional MolMiner ranges from 1.01$\times$ on the saturated drug-like filter (where the unconditional baseline is already near ceiling) to 5.25$\times$ on 3D-rich-rigid molecules, with the strongest lifts on tight multi-property targets. The 3D-rich-rigid result is particularly notable: unconditional MolMiner under-represents this region (10.9\% vs.\ 16.2\% in the training distribution) and GDSS achieves its only baseline advantage here (31.2\%), yet conditioning lifts the hit rate to 57.5\% --- a 5.25$\times$ enrichment over unconditional MolMiner and a 3.5$\times$ enrichment over the training distribution itself. This demonstrates that prompted conditional generation can override the model's intrinsic biases, exceeding the training-set frequency on regions the model otherwise under-represents.

The selectivity heatmap (Figure~\ref{fig:selectivity_heatmap}) further shows that conditioning produces a diagonally-dominant pattern of compliance, with off-diagonal cells reflecting chemically coherent property correlations rather than independent per-filter optimization. For example, conditioning on F6 (easy-synth) suppresses F5 (3D-rich-rigid) compliance to 3.7\%, reflecting the tension between sp$^3$ saturation and synthetic accessibility; conversely, F3 (fragment) prompts boost F4 (CNS) compliance to 89.8\%. These coherent shifts indicate that conditioning operates on the joint property distribution rather than each axis in isolation. Together, these results indicate that MolMiner's multi-property conditioning translates into operationally relevant control across diverse drug-discovery-aligned property windows.

\section{Conclusion}
We presented MolMiner, a fragment-based, geometry-aware, and order-agnostic autoregressive model for inverse molecular design, incorporating strong inductive biases that improve generalization and enable reliable conditional generation. MolMiner generates molecules via chemically meaningful fragment attachments with enforced validity, while dynamically incorporating three-dimensional geometry to condition decisions on physically plausible intermediate structures. The model supports multi-property conditioning over twelve physicochemical and structural targets, enabling targeted exploration from partial user specifications.

MolMiner unifies three ingredients: (i) symmetry-aware attachment standardization for consistent fragment connections, (ii) an order-agnostic likelihood factorization that effectively averages over construction histories and provides data augmentation through diverse rollouts, and (iii) scalable high-dimensional conditioning via a GMM prior that completes unspecified properties by sampling from the empirical joint distribution.

MolMiner achieves competitive unconditional performance and well-calibrated conditional control for most properties across a wide dynamic range. When prompted with multi-property targets drawn from targeted chemistry windows, conditioning lifts hit rates by up to $5.25\times$ over the unconditional baseline and $3.5\times$ over the training distribution itself, demonstrating that conditioning can override the model's intrinsic biases by exceeding the training distribution on regions it under-represents, and providing operationally relevant control across drug-discovery-aligned regions of chemical space. Importantly, controllability emerges implicitly—without auxiliary property-specific losses or oracle evaluations during generation.

Finally, MolMiner’s stepwise fragment assembly supports interpretability and opens a path toward human--computer co-design through intervention during rollout (\emph{e.g.}, constraining fragments or attachment sites). Overall, MolMiner provides a practical foundation for controllable, interpretable molecular generation and future interactive design workflows.

\medskip
\textbf{Acknowledgements} \par 
The authors acknowledge financial support from Technical University of Denmark (DTU) through the Alliance Ph.D. scholarship, from the Independent Research Foundation Denmark (0217-00326B), the Pioneer Center for Accelerating P2X Materials Discovery (CAPeX), DNRF grant number P3, the Novo Nordisk Foundation through the Centre for Basic Machine Learning Research in Life Science (MLLS, NNF20OC0062606) Novo Nordisk Foundation (NNF20OC0065611) and Independent Research Fund Denmark (9131-00082B).

\section*{Conflict of Interest}
The authors declare no conflict of interest.

\section*{Data Availability Statement}
The data that support the findings of this study are openly available on GitHub at \url{https://github.com/raulorteg/molminer}~\cite{molminerrepo}. This includes the ZINC subset with computed properties, dataset splits, training scripts, model checkpoints, and evaluation tools required to reproduce all experiments.

\medskip
\textbf{Computational Requirements} \par
All models in this work were trained using PyTorch 2.5.0 on a single NVIDIA RTX3090. Training the final model took 1 day 16h, or 50 epochs, using a batch size of 256, and RAM usage of 31 GB. We note that these requirements reflect a research-oriented implementation. In particular, the dominant computational bottleneck arises from the data ingestion pipeline, which relies on disk-based I/O for loading precomputed data into memory. We therefore expect substantial reductions in runtime and memory usage to be achievable through standard engineering optimizations.

\medskip

\bibliographystyle{unsrtnat}
\bibliography{bibliography}

\clearpage
\renewcommand{\thesection}{S\arabic{section}}
\renewcommand{\thefigure}{S\arabic{figure}}
\renewcommand{\thetable}{S\arabic{table}}
\renewcommand{\theequation}{S\arabic{equation}}
\setcounter{section}{0}
\setcounter{figure}{0}
\setcounter{table}{0}
\setcounter{equation}{0}

\begin{center}
{\Large\textbf{Supporting Information}}\\[1ex]
{\large MolMiner: Toward Controllable, 3D-Aware, Fragment-Based Molecular Design}\\[1ex]
\end{center}
\section{Calculated properties for controlled generation}
\label{sup:dataset-stats}

We included twelve annotated molecular properties for the compounds in the dataset calculated using RDKit version 2024.3.5, whose statistics for the dataset used in this work are summarized in Table \ref{tab:dset-property-statistics}.
\begin{itemize}
    \item \textbf{logP}: Logarithm of water partition coefficient, used as a measure of lipophilicity.
    
    \item \textbf{QED (Quantitative Estimate of Drug-likeness)}: A metric reflecting the drug-likeness of a molecule.
    
    \item \textbf{SAS (Synthetic Accessibility Score)}: An empirical measure of how difficult a molecule is to synthesize ranging from 1-10. Lower values suggest simpler, more synthesizable structures.
    
    \item \textbf{FractionCSP3}: The fraction of carbon atoms in sp\textsuperscript{3} hybridization, used to quantify molecular complexity and three-dimensionality.
    
    \item \textbf{molWt (Molecular Weight)}: The total weight of a molecule.
    
    \item \textbf{TPSA (Topological Polar Surface Area)}: The surface area associated with polar atoms, influencing solubility.
    
    \item \textbf{MR (Molar Refractivity)}: A descriptor related to the molecular volume and polarizability.
    
    \item \textbf{hbd (Hydrogen Bond Donors)}: The number of hydrogen bond donors.
    
    \item \textbf{hba (Hydrogen Bond Acceptors)}: The number of hydrogen bond acceptor atoms.
    
    \item \textbf{num\_rings}: The total number of ring structures present in the molecule, which contributes to rigidity.
    
    \item \textbf{num\_rotatable\_bonds}: The count of rotatable single bonds, a measure of molecular flexibility.
    
    \item \textbf{num\_chiral\_centers}: The number of chiral centers in the molecule, indicating stereochemical complexity.
\end{itemize}

\begin{table}[hbt]
\centering
  \caption{ZINC subset statistics for the twelve conditioned molecular properties.}
  \label{tab:dset-property-statistics}
    \begin{tabular}{lrrrr}
    \toprule
    \textbf{Name} & \textbf{Mean} & \textbf{Std} & \textbf{Min} & \textbf{Max} \\
    \midrule
    logP & 2.447 & 1.448 & -6.876 & 8.252 \\
    qed & 0.736 & 0.135 & 0.112 & 0.948 \\
    SAS & 3.071 & 0.864 & 1.133 & 7.289 \\
    FractionCSP3 & 0.425 & 0.226 & 0.000 & 1.000 \\
    molWt & 328.291 & 62.029 & 150.130 & 499.998 \\
    TPSA & 63.173 & 23.033 & 0.000 & 149.700 \\
    MR & 87.957 & 17.028 & 17.490 & 151.271 \\
    hbd & 1.286 & 0.891 & 0.000 & 6.000 \\
    hba & 3.676 & 1.575 & 0.000 & 11.000 \\
    num\_rings & 2.627 & 0.989 & 0.000 & 9.000 \\
    num\_rotatable\_bonds & 4.542 & 1.561 & 0.000 & 11.000 \\
    num\_chiral\_centers & 0.956 & 0.993 & 0.000 & 11.000 \\
    \bottomrule
    \end{tabular}
\begin{minipage}{\textwidth}
\end{minipage}
\end{table}

\section{Conditionally sampling Gaussian Mixture Models}
\label{sup:sampling-gmm}
A trained conditional model supports design control of $d$ properties $\Vec{x} = (x_{1}, x_{2}, ..., x_{d})$. In practice, specifying all $d$ properties is not always possible or convenient. We aim to allow a user to control any subset of controlled properties, which we call 'observed' $\Vec{x}_{obs}$, and sample the 'missing' properties $\Vec{x}_{miss}$ from the conditional distribution $p(\Vec{x}_{miss} | \Vec{x}_{obs})$. Then, the fully reconstructed array of properties $\Vec{x} = (\Vec{x}_{obs}, \Vec{x}_{miss})$ can be used as input for the conditional model.\\
In order to sample all possible combinations of 'missing' properties given any set of possible 'observed' properties it is useful to use a GMM so the conditional distributions are easy to compute. We therefore assume that the $d$ dimensional tuples of control properties for molecules in our dataset are sampled from an unknown distribution that can be approximated by a mixture of finite gaussian distributions with unknown parameters, which are optimized in our case using Expectation Maximization.
\begin{equation}
    \Vec{x} \sim f(\Vec{x}) = \sum_{k=1}^{K}\pi_{k} \ \mathcal{N}(\Vec{x} | \mu_{k},\Sigma_{k})
\end{equation}
Then the conditional density:
\begin{equation*}
    f(\Vec{x}_{miss} | \Vec{x}_{obs}) = \frac{f(\Vec{x}_{obs} , \Vec{x}_{miss})}{f(\Vec{x}_{obs})} = \sum_{k=1}^{K}\frac{\pi_{k} \ \mathcal{N}(\Vec{x} | \mu_{k},\Sigma_{k})}{f(\Vec{x}_{obs})} = 
\end{equation*}
\begin{equation*}
     = 
    \sum_{k=1}^{K}\frac{\pi_{k} \ \mathcal{N}(\Vec{x}_{obs} | \mu_{k,obs},\Sigma_{k,obs\ obs}) \ \mathcal{N}(\Vec{x}_{miss} | \mu_{k,miss | obs},\Sigma_{k,miss | obs})}{f(\Vec{x}_{obs})}
\end{equation*}
\begin{equation*}
     = 
    \sum_{k=1}^{K}\frac{\pi_{k} \ \mathcal{N}(\Vec{x}_{obs} | \mu_{k,obs},\Sigma_{k,obs\ obs}) \ \mathcal{N}(\Vec{x}_{miss} | \mu_{k,miss | obs},\Sigma_{k,miss | obs})}{\sum_{l=1}^{K}\pi_{l} \ \mathcal{N}(\Vec{x}_{obs} | \mu_{l,obs},\Sigma_{l,obs\ obs})}
\end{equation*}
Where we used:
\begin{equation*}
     \mathcal{N}(\Vec{x} | \mu_{k},\Sigma_{k}) = \mathcal{N}(\Vec{x}_{obs} | \mu_{k,obs},\Sigma_{k,obs\ obs}) \ \mathcal{N}(\Vec{x}_{miss} | \mu_{k,miss | obs},\Sigma_{k,miss | obs})
\end{equation*}
\begin{equation*}
     f(\Vec{x}_{obs}) = \sum_{l=1}^{K}\pi_{l} \ \mathcal{N}(\Vec{x}_{obs} | \mu_{l,obs},\Sigma_{l,obs\ obs})
\end{equation*}
Re-organizing the conditional density we write:
\begin{equation}
    f(\Vec{x}_{miss} | \Vec{x}_{obs}) = 
    \sum_{k=1}^{K} w_{k} \ \mathcal{N}(\Vec{x}_{miss} | \mu_{k,miss | obs},\Sigma_{k,miss | obs})
\end{equation}
With:
\begin{equation*}
    w_{k} =  \frac{\pi_{k} \ \mathcal{N}(\Vec{x}_{obs} | \mu_{k,obs},\Sigma_{k,obs\ obs}) }{\sum_{l=1}^{K}\pi_{l} \ \mathcal{N}(\Vec{x}_{obs} | \mu_{l,obs},\Sigma_{l,obs\ obs})}
\end{equation*}
\begin{equation*}
    \mu_{k,miss | obs} = \mu_{k,miss} + \Sigma_{k,miss\ obs}\Sigma_{k,obs\ obs}^{-1} (\Vec{x}_{obs}-\mu_{k,obs})
\end{equation*}
\begin{equation*}
    \Sigma_{k,miss | obs} = \Sigma_{k,miss \ miss} - \Sigma_{k,miss \ obs}\Sigma_{k,obs \ obs}^{-1}\Sigma_{k,obs \ miss}
\end{equation*}
Given $\Vec{x}_{obs}$ sampling the conditional distributions can then be done first sampling the k-th gaussian to use, weighted by the scalar $w_{k}$ and then sampling the corresponding gaussian mixture,
\[
\Vec{x}_{miss} \sim \mathcal{N}(\mu_{k,miss | obs},\Sigma_{k,miss | obs}).
\]

\section*{Implementation \& Hyperparameters}
\begin{figure}[ht]
    \centering
    \includegraphics[width=0.6\linewidth]{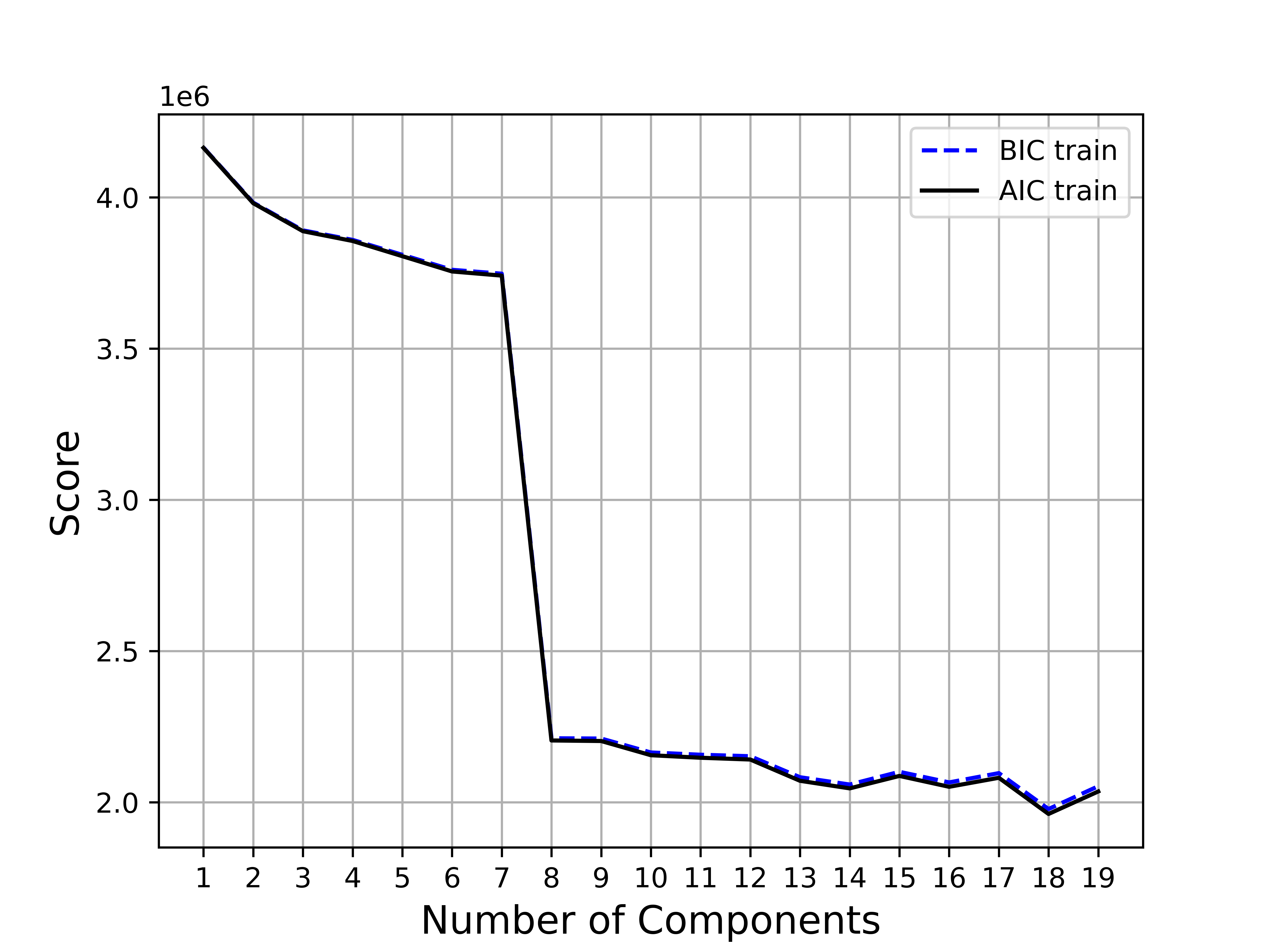}
    \caption{Elbow visualization of the BIC and AIC scores for varying number of GMM components ranging 1-19. Note that at K=8 there is a sharp drop, elbow, which marks an ideal number of components to use for this problem.}
    \label{fig:gmm-elbow}
\end{figure}

We employed the GMM implementation from scikit-learn (version 1.5.2), specifically the \textit{GaussianMixture} class. In order to find the optimal number of components K of the mixture we employ the elbow method with BIC and AIC metrics, as implemented in the aforementioned python package. The GMM model was trained using the same dataset split train-val-test splits (80-10-10) as the rest of this work. For varying number of mixture components the models were trained on the training set and the BIC and AIC scores noted, as shown in Figure \ref{fig:gmm-elbow}.

From Figure \ref{fig:gmm-elbow} we take the K=8 as the best number of components for the GMM in our problem as it is located in the elbow of the plot, indicating an ideal trade-off between complexity and fidelity.

\section*{Further validation}
\begin{figure}[ht]
    \centering
    \includegraphics[width=\linewidth]{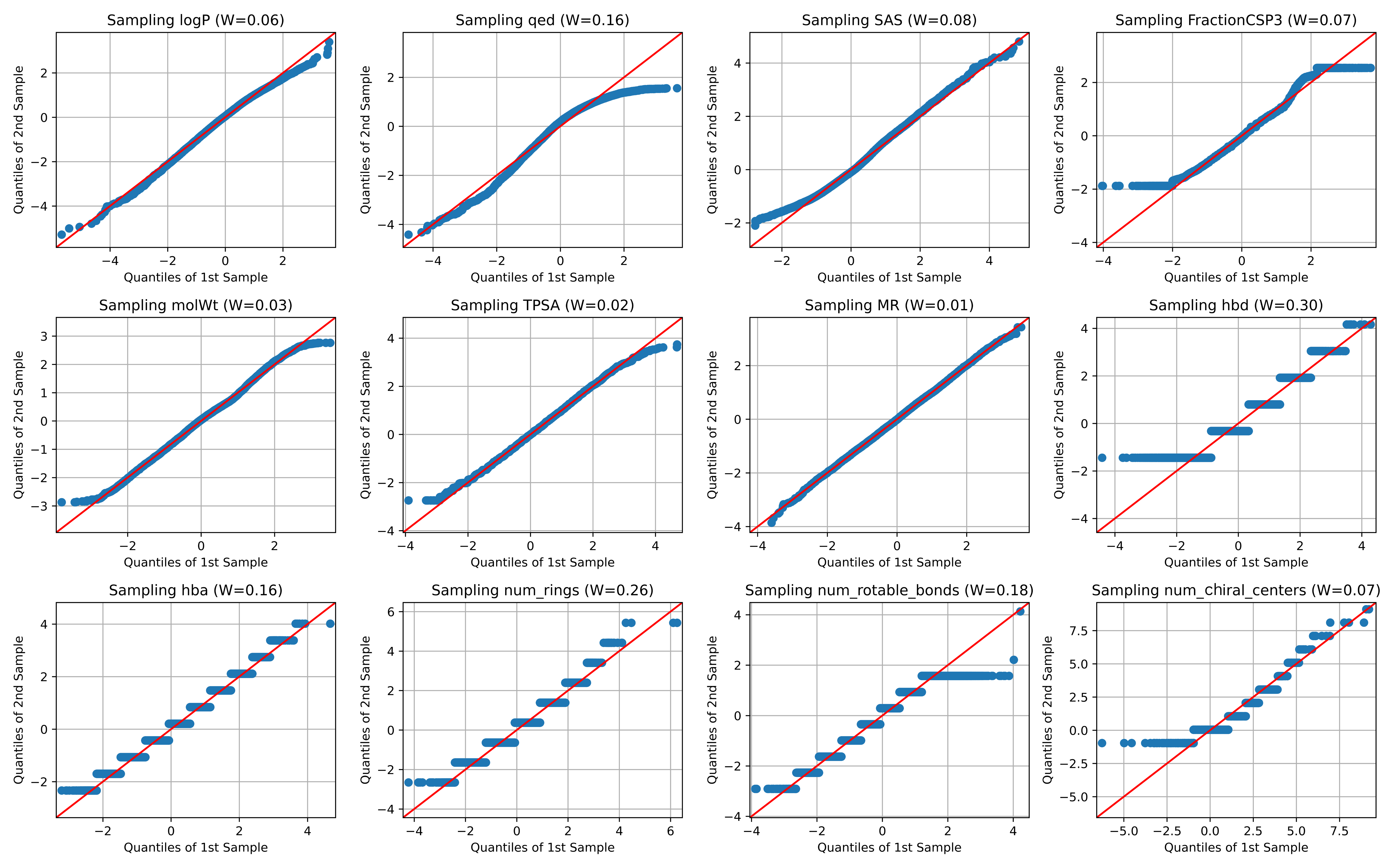}
    \caption{Reconstruction fidelity of 1 missing property given all others. Using the validation dataset, for every property the model is asked to reconstruct one of the properties given the rest, then the reconstructed and real distributions are compared using q-q plots annotated with the Wasserstein (W) distance.}
    \label{fig:gmm-qqplots}
\end{figure}

Using 8 number of components, we next perform an experiment to evaluate how well does the GMM capture the underlying multi-variate distribution using the validation dataset. 
We evaluate how good the model is at reconstructing one property given all others. For each property, we mask it and ask the model to reconstruct it, then we produce the quantile-quantile (q-q) plots and compute the Wasserstein (W) distance between the real values and those reconstructed. The result of this first experiment are shown in Figure \ref{fig:gmm-qqplots}.

From Figure \ref{fig:gmm-qqplots}, we can validate that the reconstruction of the missing property using the GMM resembles the true underlying conditional distribution, the quantile-quantile plots fall mostly along the 45 degrees ideal line, and the Wasserstein distances are small, further indicating good agreement. Note that for properties such as 'number of rings' the q-q plots show horizontal lines as a result of treating the discrete values as resulting from a continuous distribution. 

\section{Ablation studies \& Hyperparameter search}
\label{sup:ablation-studies}
We performed a hyperparameter sweep over learning rate ($3\text{e-}5$, $5\text{e-}5$), dropout (0.1–0.3), warmup ratio (0.15, 0.2), and number of attention heads (16, 64, 128), selecting $5\text{e-}5$ peak LR, 0.15 warmup, 0.3 dropout, and 16 heads as the best configuration.

\subsubsection*{The effect of geometry}
\begin{figure}[H]
    \centering
    \includegraphics[width=0.8\linewidth]{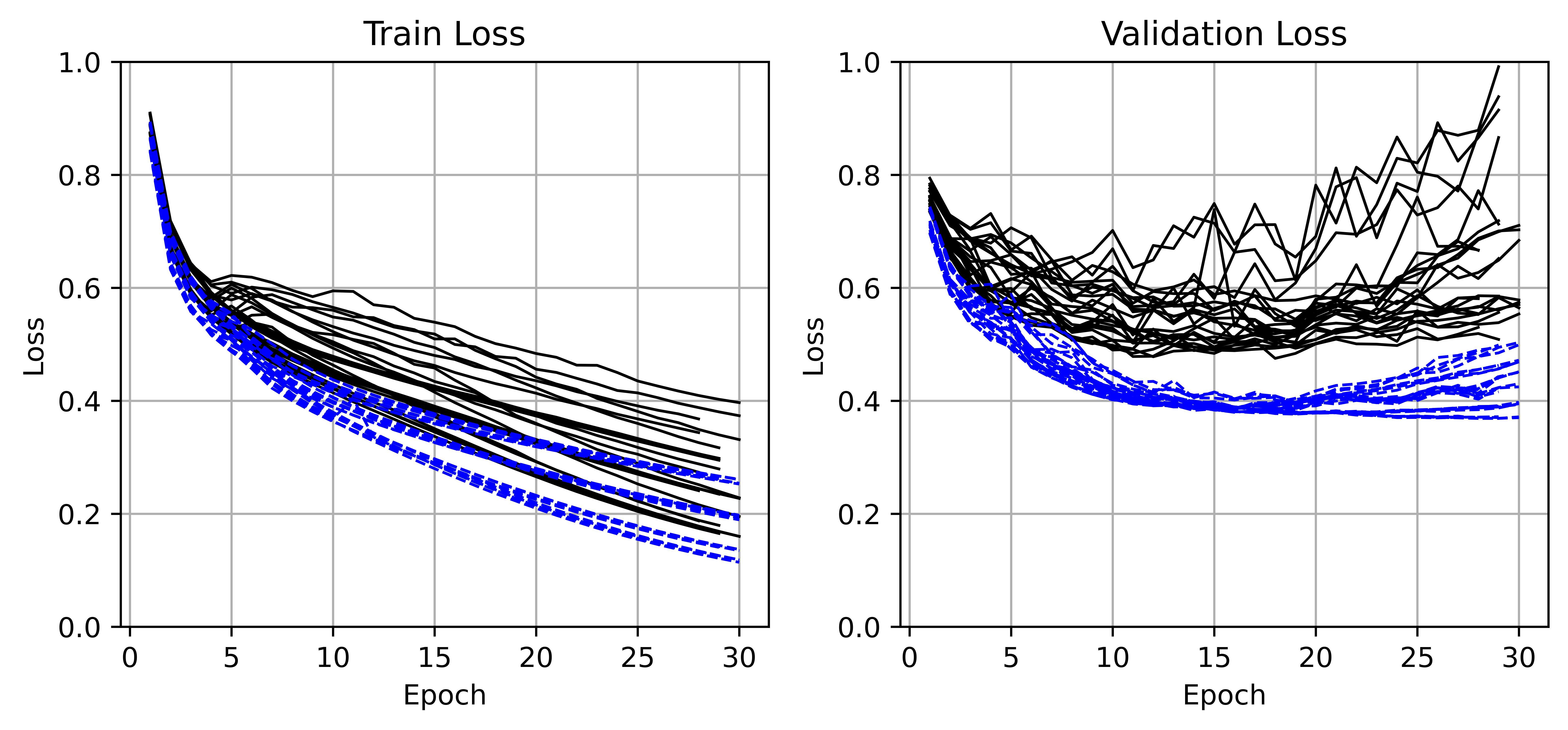}
    \caption{Training and validation curves for models with and without geometric information. (black) models where the geometric weighting factor was fixed at zero and not optimized. (blue) models with geometric factor initialized at one and optimized. Models incorporating geometric information (not null, blue) consistently outperform those without, highlighting the importance of geometry in model performance.}
    \label{fig:train-curve-geometry}
\end{figure}

\begin{figure}[H]
    \centering
    \includegraphics[width=0.8\linewidth]{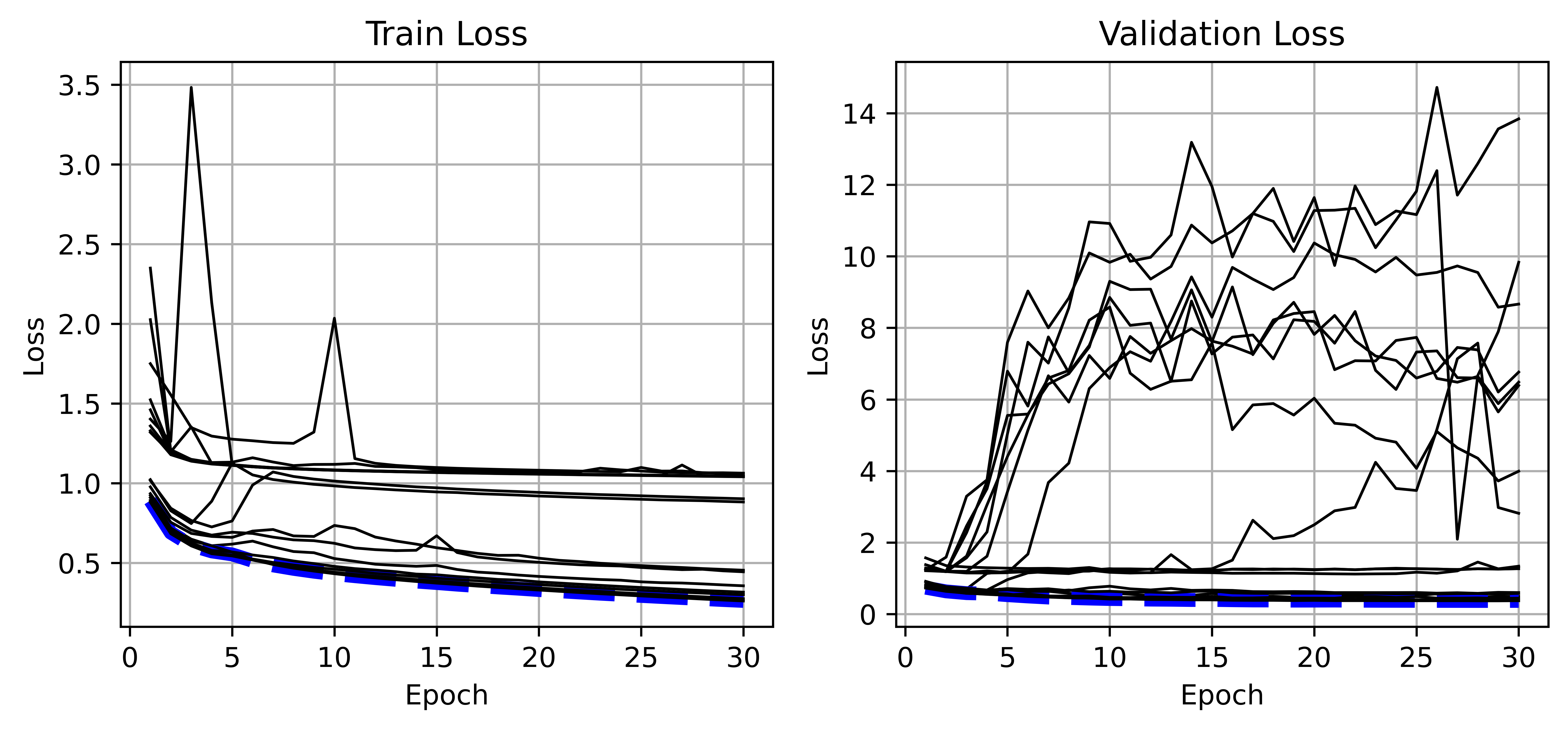}
    \caption{Training and validation curves for models with trainable geometric weighting factor, initialized with values ranging from ±1000 to 0. The best-performing model (blue) corresponds to initialization +1. Models initialized with large negative values diverge during training—consistent with the incorrect weighting of geometry where distant atoms are emphasized more than nearby ones.}
    \label{fig:train-curve-geomsearch}
\end{figure}

We performed a two-part study to evaluate the influence of geometric information on model performance. First, we conducted an ablation study by comparing models trained with geometry (global geometry attention bias factor initialized at 1 and trained) versus without geometry (factor fixed at 0). As shown in Figure~\ref{fig:train-curve-geometry}, incorporating geometry consistently improved both training and validation reconstruction loss. We then performed a hyperparameter search, initializing the geometry factor across a wide range (from large negative to large positive values) while allowing it to be trained (Figure~\ref{fig:train-curve-geomsearch}). Models initialized with large negative values—which emphasize distant over proximal atomic relationships—exhibited poor performance or training instability. The best results were obtained with moderate positive initializations, with +1 yielding the strongest performance.

\section{Hyperparameter search for SeedFragNet model}
\begin{figure}[ht]
    \centering
    \includegraphics[width=0.8\linewidth]{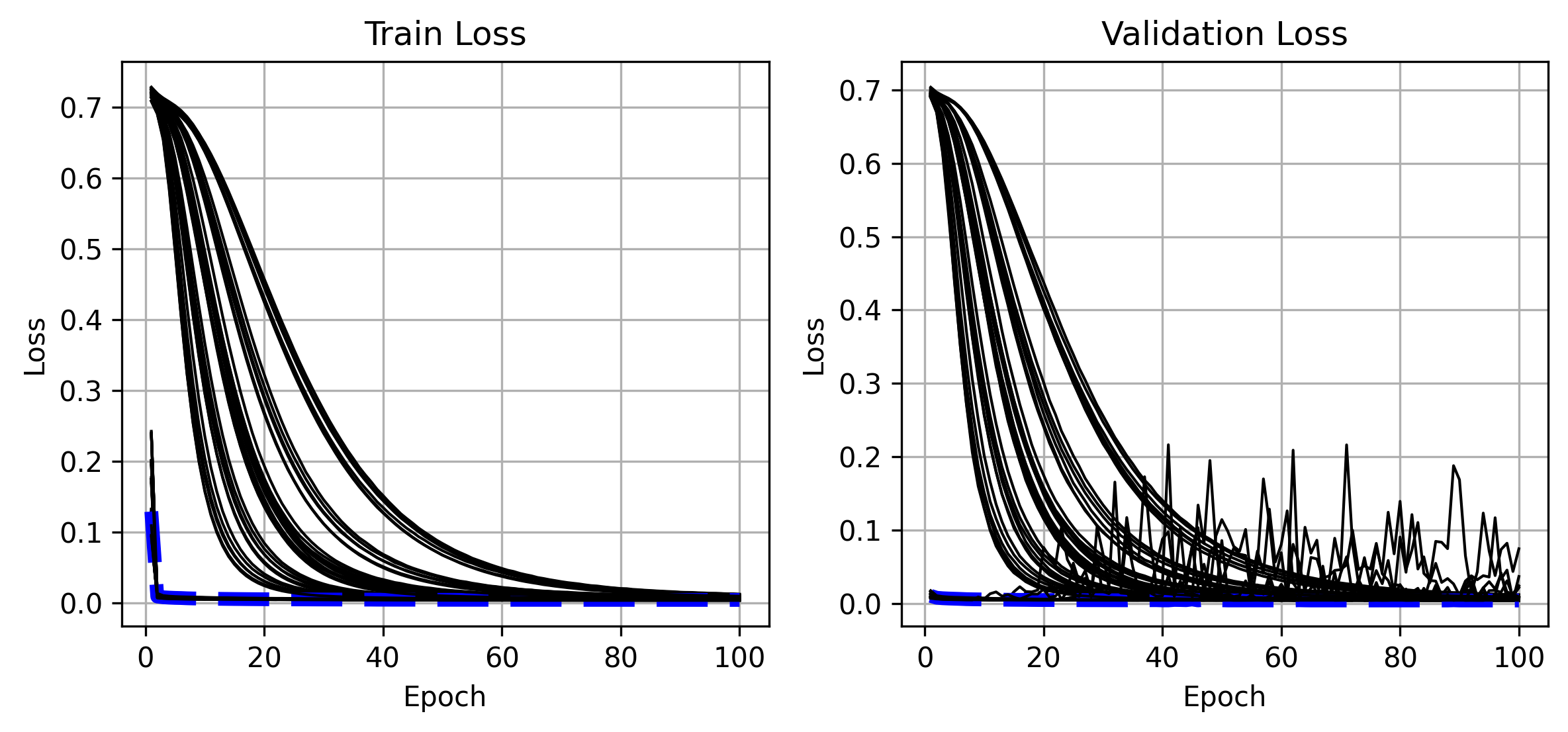}
    \caption{Training and validation loss curves for all hyperparameter configurations. Each line represents a separate model run. The selected final model, chosen based on the lowest validation loss, is highlighted in blue. Notably, several models — including the selected configuration — exhibit substantially lower initial loss values. We hypothesize this is due to favorable random initialization.}
    \label{fig:train-curve-starter}
\end{figure}

We conducted a grid search over the following hyperparameters: dff (number of neurons per layer) [512, 1024, 2058], batch sizes [512, 1024], number of layers [2, 3, 4], dropout probabilities [0.1, 0.3, 0.5], and learning rate factors [0.5, 0.8]. Each model was trained for 100 epochs using an initial learning rate of 1e-4 and a fixed seed of 42, resulting in a total of 108 model configurations. The best-performing model was selected based on validation set performance. Figure \ref{fig:train-curve-starter} presents the training and validation loss curves for all runs, with the final selected model highlighted in blue. This model used dff = 512, batch size = 512, 3 layers, a dropout probability of 0.5 and lr factor 0.5. We used nn.BCEWithLogitsLoss() as the loss function. The training produced a mean training loss of 0.0052 $\pm$ 0.0001 and a mean validation loss of 0.0047 $\pm$ 0.0001. 

\section{Experiment: Influence of Conditioning Properties on Initial Fragment Selection}
\label{sup:seed-flow}
\begin{figure}[ht]
    \centering
    \includegraphics[width=\linewidth]{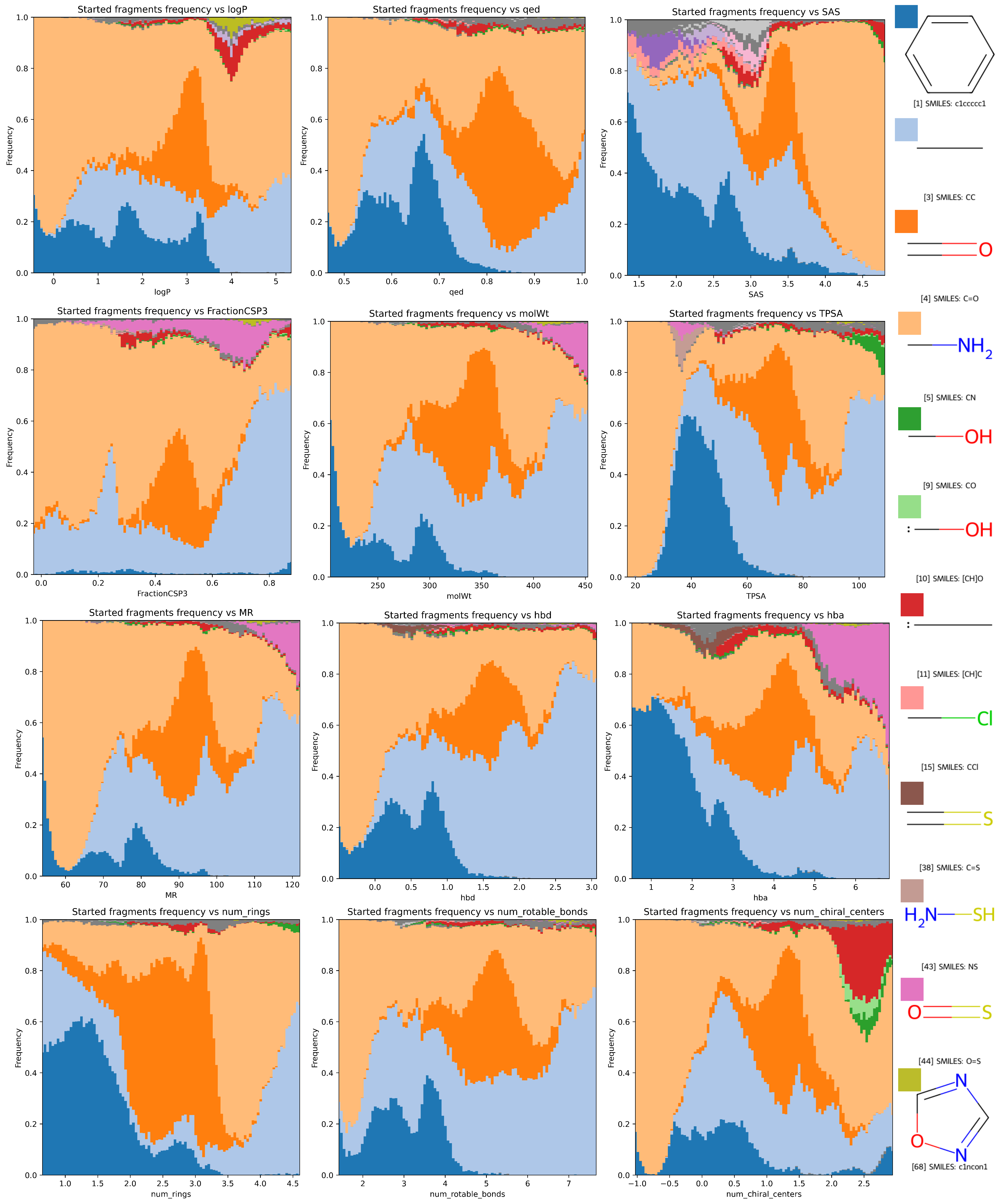}
    \caption{Effect of conditioning properties on fragment selection. Each subplot corresponds to one of the 12 molecular properties used for conditioning. The resulting flow plots show the frequency distribution of selected fragments as a function of the conditioned property, revealing how different properties influence the initial fragment choice during generation.}
    \label{fig:fragmentstarter-flow}
\end{figure}

To investigate how individual conditioning properties affect the seed fragment selection process in our molecular generator, we conducted a controlled experiment. The SeedFragNet model, which selects an initial fragment based on a tuple of 12 conditioning properties, was used to generate fragments while systematically varying one property at a time. For each property, we sampled 100 evenly spaced values within [$-2\sigma$, $+2\sigma$] and, for each value, generated 100 random samples of the remaining 11 properties using a GMM fitted to the training data. These condition vectors were fed into the SeedFragNet model, and the resulting fragment selections were recorded. The outcome, visualized as flow plots (Figure \ref{fig:fragmentstarter-flow}), shows how the frequency of fragment choices evolves as a function of each conditioning property, offering insight into the dependency between the model’s conditioning inputs and seed fragment selection behavior.

\clearpage

\section{Fragment decomposition and tokenization of molecular graphs}
\label{sup:coarse-graining}
We represent molecules as assemblies of non-overlapping, interpretable fragments. To construct these, we apply a coarse-graining strategy that systematically decomposes each molecule into a set of irreducible substructures. First, we identify all rings using the SSSR algorithm, which captures the fundamental cyclic motifs of the molecule. Any remaining bonds, connecting atoms not assigned to a ring, are treated as simple two-atom fragments.

This decomposition follows three key principles:
First, it ensures \textbf{uniqueness}: the same molecule will always be partitioned into the same set of irreducible fragments, independent of atom ordering or indexing.
Second, it guarantees \textbf{disjointness}, so no two fragments share atoms (other than at their attachment points), allowing unambiguous reconstruction of the original molecule.
Finally, it emphasizes \textbf{interpretability}: each fragment corresponds to a recognizable unit, such as a ring or bond, promoting alignment between the model’s representation and chemical intuition.

While alternative strategies could use predefined functional groups as fragments, this introduces ambiguity. Functional groups are often composed of smaller substructures that may also occur independently, raising the question of whether to treat the group as a single fragment or to decompose it further. Our approach avoids this ambiguity by consistently decomposing molecules into irreducible fragments.

Additionally, our choice leads to a desirable property: every fragment corresponds to a single cycle—either a ring or a two-atom ``bond cycle.'' This simplifies the identification of symmetries, since canonicalization and index remapping reduce to cyclic permutations, which are straightforward to compute and resolve.

\paragraph{Fragment extraction code.} The following function implements the fragment extraction procedure:

\begin{verbatim}
def _find_clusters(self, molecule) -> list:
    """
    Given an RDKit molecule object, find the list of lists 
    containing atom indexes belonging to each grain.
    """
    clusters = [list(x) for x in Chem.rdmolops.GetSSSR(molecule)]
    for bond in molecule.GetBonds():
        a1 = bond.GetBeginAtom().GetIdx()
        a2 = bond.GetEndAtom().GetIdx()
        bond_in_existing_clusters = False
        for cluster in clusters:
            if (a1 in cluster) and (a2 in cluster):
                bond_in_existing_clusters = True
                break
        if not bond_in_existing_clusters:
            clusters.append((a1, a2))
    return clusters
\end{verbatim}

\begin{figure}[ht]
    \centering
    \includegraphics[width=\linewidth]{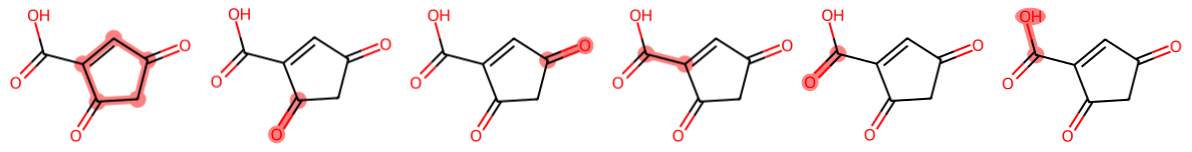}
    \caption{Extracted fragments highlighted on the molecular graph for \texttt{O=C1CC(=O)C=C1C(=O)O}.}
    \label{fig:highlighted-grains}
\end{figure}

Each fragment is represented by its \textit{Canonical SMILES} (RDKit), providing a compact, indexing-invariant encoding that is both human- and machine-readable. An example of extracted fragments is shown in Figure~\ref{fig:highlighted-grains}.

\subsubsection*{Tracking Attachment Points}
\begin{figure}[ht]
    \centering
    \includegraphics[width=\linewidth]{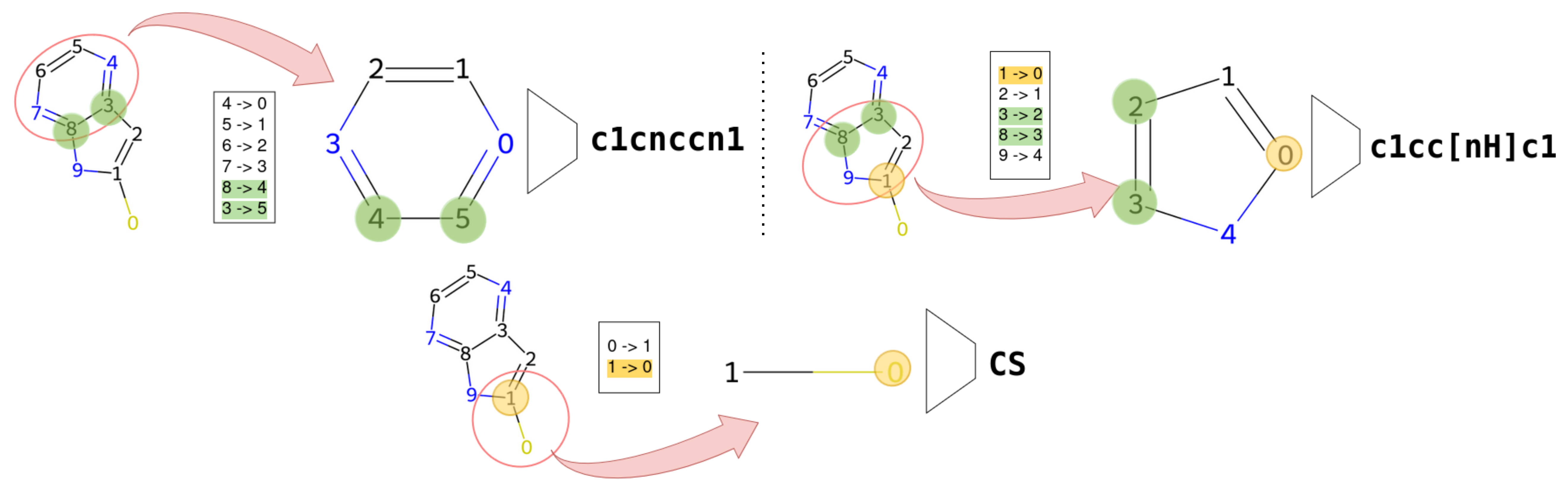}
    \caption{Fragment extraction and attachment point tracking. Atoms shared between fragments define attachment points (highlighted in green).}
    \label{fig:fragment-attachment}
\end{figure}

While Canonical SMILES provide a convenient representation, they omit explicit attachment information — that is, how fragments connect within the original molecule. To address this, we maintain a mapping between the \textit{global atom indices} (corresponding to the original molecule) and the \textit{local indices} (assigned within each extracted fragment). Atoms shared between fragments naturally define the attachment points, as illustrated in Figure~\ref{fig:fragment-attachment}.

\subsubsection*{Canonicalization and Index Permutations}
\begin{figure}[ht]
    \centering
    \includegraphics[width=\linewidth]{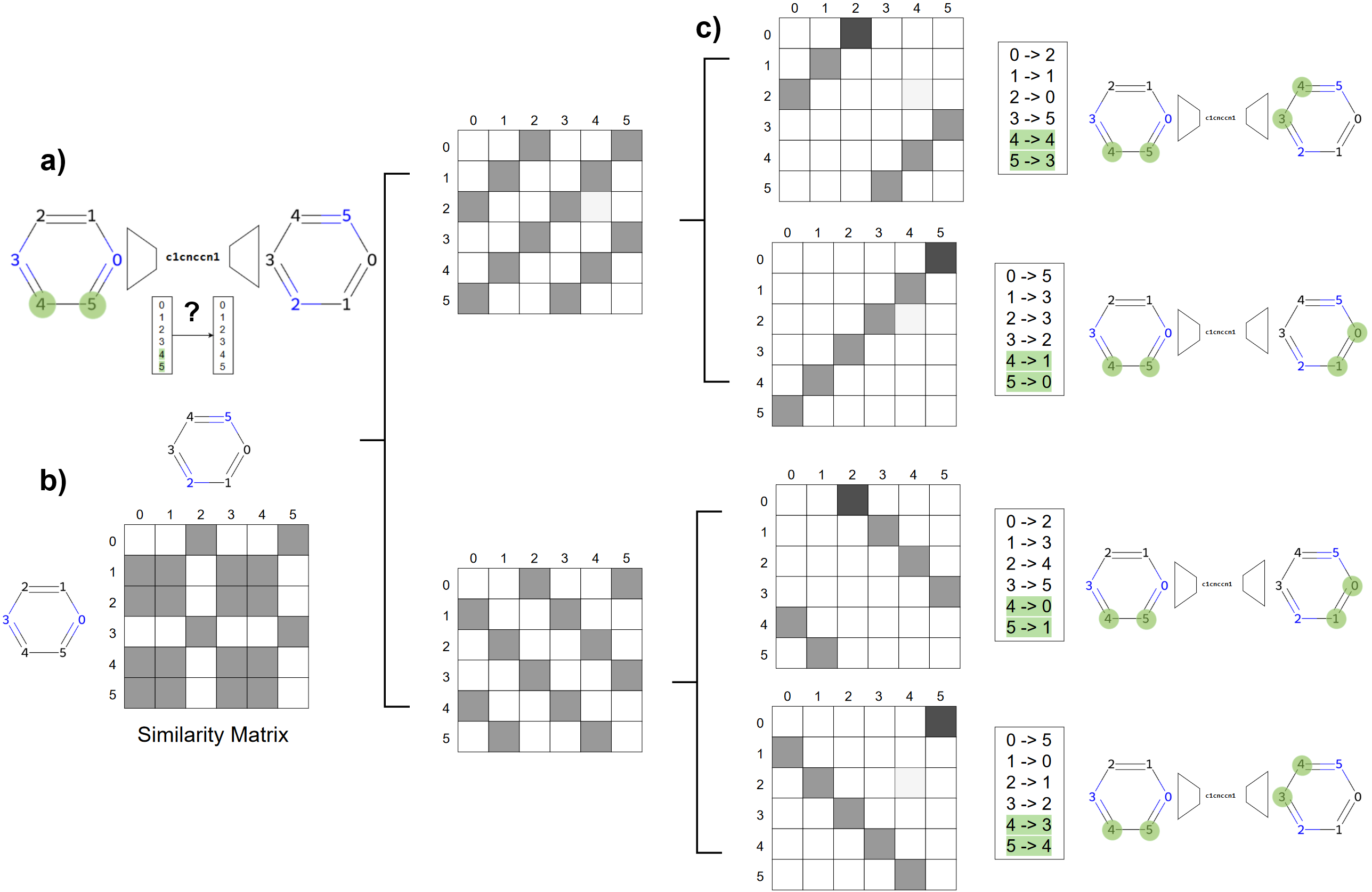}
    \caption{Extraction of all possible maps between local atom indices.
    (\textbf{a}) When a fragment is canonicalized and later reconstructed, its local atom indices may change, resulting in an unknown correspondence between the original and new indices.
    (\textbf{b}) To recover this mapping, we compute the Tanimoto similarity between Morgan fingerprints of every pair of atoms, yielding a similarity matrix.
    (\textbf{c}) Given the prior knowledge that reindexing must correspond to a cyclic permutation, we identify all cyclic shifts that preserve high-similarity correspondences, extracting the valid mappings between the original and new local indices.}
    \label{fig:fragment-mapping}
\end{figure}

\begin{figure}[ht]
\centering
\includegraphics[width=\linewidth]{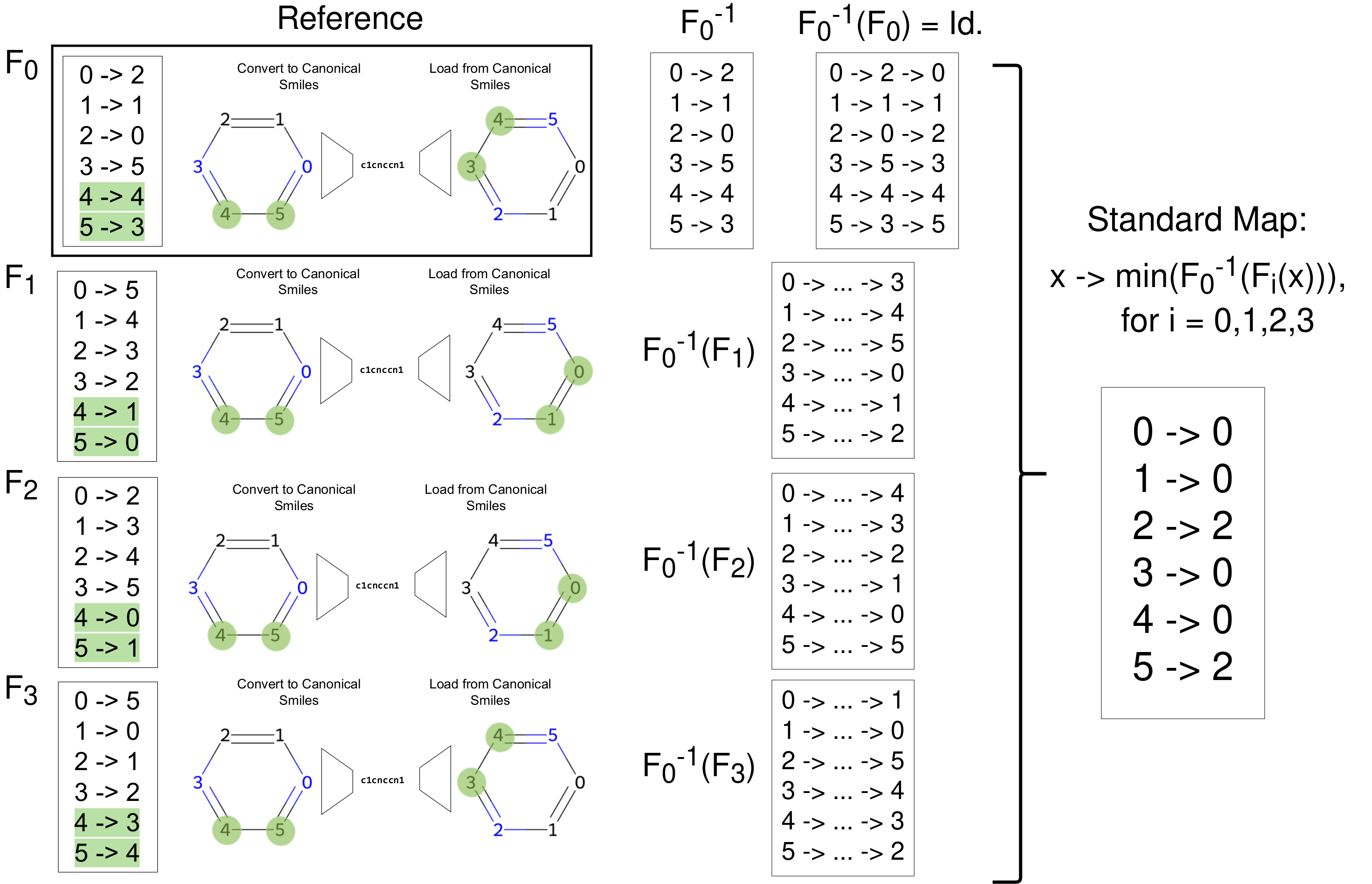}
\caption{Extraction of fragments and attachment points from a molecular graph, followed by the construction of the standard map to resolve index correspondences in a canonical, symmetry-invariant manner.}
\label{fig:extracted-grains-2}
\end{figure}

When a fragment is re-created from its Canonical SMILES, the local atom indices may change due to the canonicalization process. Fortunately, because every fragment corresponds to a single cycle (either a ring or a two-atom bond cycle), this reindexing is always equivalent to a cyclic permutation of the original atom ordering.
To recover the index mapping between the original fragment and its canonicalized form, we compute pairwise atom similarities using the Tanimoto similarity of Morgan fingerprints, as detailed in Figure~\ref{fig:fragment-mapping}. This produces a similarity matrix, from which we identify cyclic shifts that yield consistent, high-similarity correspondences.
To achieve consistent and symmetry-invariant identification of attachment points across fragments, we construct a map that resolves all permissible index correspondences into a reference form. The procedure is illustrated in Figure \ref{fig:extracted-grains-2}.
Let $\{F_i\}$ denote the set of all valid index mappings between the original fragment atom indices and those assigned upon canonicalization, where $i = 0, 1, \dots, n$. We select the first mapping, $F_0$, as the reference and compute its inverse $F_0^{-1}$. For each mapping $F_i$, we then construct a composite mapping defined as:
\begin{equation}
M_i(x) = F_0^{-1}(F_i(x)).
\end{equation}
This composition aligns all mappings relative to the reference indexing. To derive the final map, for each atom index $x$, we identify the minimal mapped value across all composite mappings:
\begin{equation}
x \longmapsto \min_{i} \left( M_i(x) \right) = \min_{i} \left( F_0^{-1}(F_i(x)) \right).
\end{equation}

This selection yields a unique, deterministic correspondence that is invariant to atom reindexing and respects molecular symmetries, including those arising from automorphisms. The resulting standard map provides a robust framework for attachment point tracking, ensuring consistency in fragment reassembly and downstream graph-based computations.

\clearpage
\section{Ablation and Sensitivity Analysis of Sampling Strategies}
\label{sup:ablation-sampling}
To identify the most effective sampling configuration for our generative model, we conducted a systematic ablation and sensitivity analysis across key components of the sampling process. While the primary goal of this study was to optimize generation quality, the experimental design also provides insight into how individual sampling choices influence the balance between validity, novelty, diversity, and uniqueness.

Our sampling pipeline includes several controllable factors: (i) the source of molecular conditions (either directly from the dataset or sampled from a learned GMM), (ii) the decoding strategy (greedy versus probabilistic sampling during autoregressive generation), (iii) the size of the candidate pool ($k$) from which the initial seed fragment is selected, and (iv) the method for choosing the seed fragment itself (random or weighted by the model’s predicted scores). Each of these decisions has the potential to impact the quality and chemical diversity of the generated molecules.

We report the results of varying these factors systematically in Table~\ref{tab:supl-wdist-property-statistics}. The findings not only reveal the most effective configuration for our task but also clarify how different sampling strategies contribute to performance trade-offs. This analysis serves both as a guide for practical deployment and as an ablation study highlighting the robustness and flexibility of our model.

\begin{table}[H]
\centering
\scriptsize
\caption{Wasserstein distances between the property distributions of generated molecules (N $\approx$ 5,000) and the reference dataset are reported, along with uniqueness (\%), novelty (\%), and mean Tanimoto similarity, across various sampling strategies. For example, S/3/NW/NG indicates that conditions were sampled from the GMM (S) rather than the dataset (D); the initial fragment was selected from the top $k=3$ candidates; selection among these was random and unweighted (NW); and subsequent fragments were sampled in a non-greedy (NG), probability-weighted manner during the autoregressive process. For each metric, the best-performing configuration is highlighted in bold. Overall, the best configuration was D/5/W/NG, while among sampled-only settings, S/10/W/NG performed best, with its termination-loss correction ``$\ast$S/10/W/NG'' improving its performance.}
\label{tab:supl-wdist-property-statistics}

{%
  \setlength{\tabcolsep}{5pt}
\begin{tabular}{lrrrrrrrrrrrrrrr}
\textbf{Config.} & 
\rotatebox{90}{logP} & 
\rotatebox{90}{QED} & 
\rotatebox{90}{SAS} & 
\rotatebox{90}{FractCSP3} & 
\rotatebox{90}{molWt} & 
\rotatebox{90}{TPSA} & 
\rotatebox{90}{MR} & 
\rotatebox{90}{HBD} & 
\rotatebox{90}{HBA} & 
\rotatebox{90}{\#Rings} & 
\rotatebox{90}{\#RotBonds} & 
\rotatebox{90}{\#Chiral} &
\rotatebox{90}{\%Uniqueness} & 
\rotatebox{90}{\%Novelty} & 
\rotatebox{90}{Diversity}\\
\hline
S/3/NW/NG & 0.53 & 0.02 & 0.08 & \textbf{0.02} & 67.23 & 10.99 & 16.99 & 0.15 & 0.59 & 0.61 & 0.92 & 0.25 & 98.0 & 99.8 & \textbf{0.89} \\
S/3/NW/G& 0.60 & 0.03 & 0.26 & 0.03 & 83.88 & 15.28 & 20.46 & 0.20 & 0.92 & 0.78 & 0.96 & 0.33 & 91.6 & 99.8 & 0.88 \\
D/3/NW/NG & 0.33 & \textbf{0.01} & \textbf{0.06} & \textbf{0.02} & 49.79 & 8.02 & 12.60 & \textbf{0.14} & 0.41 & 0.42 & 0.73 & 0.20  & 98.7 & 99.3 & 0.88 \\
D/3/NW/G & 0.44 & \textbf{0.01} & 0.20 & 0.04 & 63.35 & 10.51 & 15.53 & 0.17 & 0.62 & 0.52 & 0.86 & 0.24 & 92.1 & 98.3 & 0.87 \\
S/3/W/NG & 0.51 & 0.02 & 0.08 & \textbf{0.02} & 67.48 & 11.75 & 17.03 & 0.16 & 0.60 & 0.59 & 0.92 & 0.26 & 97.3 & 99.9 & \textbf{0.89} \\
S/3/W/G & 0.62 & 0.03 & 0.28 & 0.03 & 84.94 & 15.38 & 20.61 & 0.21 & 0.91 & 0.80 & 0.96 & 0.34 & 90.7 & 99.8 & 0.88 \\
D/3/W/NG & 0.34 & \textbf{0.01} & 0.07 & \textbf{0.02} & 49.00 & 7.76 & 12.39 & \textbf{0.14} & \textbf{0.36} & 0.41 & 0.75 & 0.21  & 98.8 & 99.7 & \textbf{0.89} \\
D/3/W/G & 0.50 & 0.02 & 0.17 & 0.03 & 66.57 & 11.22 & 16.45 & 0.17 & 0.66 & 0.56 & 0.91 & 0.26 & 90.9 & 98.4 & 0.87 \\
S/5/NW/NG & 0.54 & 0.02 & 0.09 & \textbf{0.02} & 66.45 & 10.87 & 16.89 & 0.16 & 0.57 & 0.61 & 0.90 & 0.24 & 97.6 & 99.9 & \textbf{0.89} \\
S/5/NW/G & 0.63 & 0.04 & 0.24 & 0.03 & 85.85 & 15.88 & 21.20 & 0.23 & 0.94 & 0.80 & 1.03 & 0.32 & 90.7 & 99.9 & 0.88 \\
D/5/NW/NG& 0.37 & \textbf{0.01} & 0.07 & \textbf{0.02} & 49.64 & 7.75 & 12.72 & 0.16 & 0.37 & 0.42 & 0.76 & 0.20 & 98.6 & 99.4 & \textbf{0.89} \\
D/5/NW/G & 0.44 & 0.02 & 0.18 & 0.04 & 65.50 & 11.47 & 16.19 & 0.18 & 0.65 & 0.55 & 0.87 & 0.23 & 91.3 & 98.5 & 0.87 \\
S/5/W/NG& 0.51 & 0.02 & 0.09 & \textbf{0.02} & 68.08 & 11.60 & 17.13 & 0.16 & 0.59 & 0.61 & 0.92 & 0.27 & 97.4 & \textbf{100.0} & \textbf{0.89} \\
S/5/W/G& 0.61 & 0.03 & 0.25 & 0.03 & 85.22 & 15.60 & 20.62 & 0.21 & 0.92 & 0.79 & 0.98 & 0.33 & 90.4 & 99.6 & 0.88 \\
\rowcolor{lightgray}\textbf{D/5/W/NG}& \textbf{0.31} & \textbf{0.01} & 0.07 & \textbf{0.02} & \textbf{47.23} & 7.59 & \textbf{11.91} & 0.14 & \textbf{0.36} & 0.41 & \textbf{0.64} & 0.19  & \textbf{98.9} & 99.5 & \textbf{0.89} \\
D/5/W/G& 0.47 & 0.02 & 0.19 & 0.03 & 67.27 & 11.95 & 16.66 & 0.21 & 0.68 & 0.55 & 0.97 & 0.26 & 91.0 & 98.5 & 0.87 \\
S/10/NW/NG& 0.58 & 0.02 & 0.10 & \textbf{0.02} & 68.67 & 10.77 & 17.70 & 0.16 & 0.55 & 0.62 & 0.96 & 0.25 & 98.2 & 99.8 & \textbf{0.89} \\
S/10/NW/G& 0.65 & 0.04 & 0.29 & 0.05 & 89.62 & 16.44 & 22.60 & 0.27 & 0.95 & 0.82 & 1.16 & 0.34 & 89.1 & 99.6 & 0.88 \\
D/10/NW/NG& 0.41 & \textbf{0.01} & 0.07 & 0.03 & 51.93 & \textbf{7.53} & 13.57 & 0.15 & \textbf{0.36} & 0.45 & 0.78 & \textbf{0.18}  & 98.7 & 99.6 & \textbf{0.89} \\
D/10/NW/G& 0.47 & 0.02 & 0.24 & 0.05 & 68.32 & 11.85 & 17.44 & 0.23 & 0.66 & 0.58 & 0.97 & 0.27 & 90.6 & 98.9 & 0.88 \\
\rowcolor{lightgray}\textbf{S/10/W/NG}& 0.46 & 0.02 & 0.09 & \textbf{0.02} & 64.67 & 10.88 & 16.27 & 0.16 & 0.56 & 0.59 & 0.88 & 0.26 & 97.6 & 99.8 & \textbf{0.89} \\
\rowcolor{lightgray}\textbf{$\ast$S/10/W/NG}& 0.38 & \textbf{0.01} & 0.09 & \textbf{0.02} & 52.09 & 8.77 & 13.54 & \textbf{0.12} & 0.45 & 0.54 & 0.70 & 0.20 & 98.8 & 99.9 & \textbf{0.89} \\
S/10/W/G& 0.63 & 0.03 & 0.25 & 0.03 & 84.00 & 14.84 & 20.40 & 0.21 & 0.87 & 0.78 & 0.96 & 0.33 & 90.7 & 99.7 & 0.88 \\
D/10/W/NG& 0.33 & \textbf{0.01} & \textbf{0.06} & \textbf{0.02} & 48.55 & 7.87 & 12.24 & 0.15 & \textbf{0.36} & \textbf{0.40} & 0.72 & 0.19  & 98.7 & 99.5 & \textbf{0.89} \\
D/10/W/G& 0.47 & 0.02 & 0.19 & 0.03 & 64.86 & 10.88 & 16.02 & 0.18 & 0.63 & 0.55 & 0.89 & 0.26 & 91.4 & 98.6 & 0.87 \\
\end{tabular}
}
\end{table}

\section{Specialist-vs-Generalist ablation}\label{sec:si-generalistvsspecialist}

To directly quantify the trade-off between high-dimensional conditioning and per-property accuracy, we trained an additional variant of MolMiner, denoted MolMiner-logP, in which all conditioning channels except logP are masked during training. This variant shares the same architecture, hyperparameters, training data, and rollout procedure as the full 12-property MolMiner, isolating conditioning dimensionality as the only difference. MolMiner-logP therefore acts as an internal specialist baseline, and the comparison between the two variants provides a clean measurement of the cost and benefit of generalist conditioning. We evaluate both variants under the same protocols as the main text: unconditional distributional fidelity (Section~\ref{sec:uncond-gen}) and per-property conditional calibration (Section~\ref{subsec:cond}).

\begin{figure}[H]
    \centering
    \includegraphics[width=\linewidth]{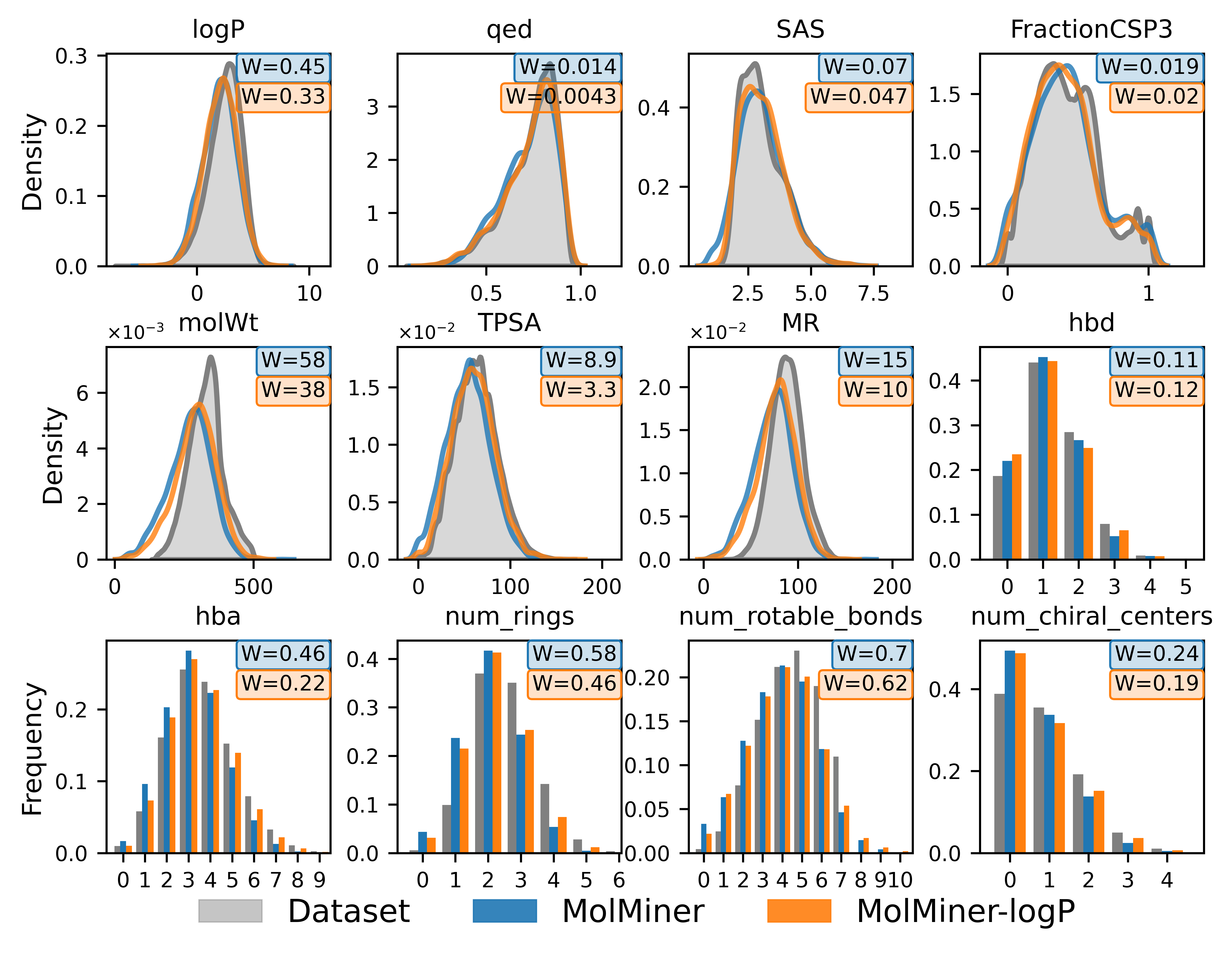}
    \caption{Kernel density estimates (KDE) and histograms of twelve molecular properties across 5{,}000 generated molecules. Distributions from the reference dataset are compared against MolMiner (12D generalist) and MolMiner-logP (1D specialist)}
    \label{fig:si-generation-unconditional-benchmark-dists}
\end{figure}

Figure~\ref{fig:si-generation-unconditional-benchmark-dists} compares the marginal property distributions of MolMiner-logP against those of the full 12-property MolMiner and the reference dataset. MolMiner-logP matches the reference marginals more closely on most properties, with the largest improvements observed on molecular weight (W: $58 \rightarrow 38$), TPSA (W: $8.9 \rightarrow 3.3$), MR (W: $15 \rightarrow 10$), and hba (W: $0.46 \rightarrow 0.22$). This behavior is expected: a model conditioned on fewer properties is asked to satisfy fewer simultaneous constraints during generation, and is therefore freer to reproduce the empirical marginals. Generalist conditioning trades a small amount of unconditional distributional fidelity for the ability to control eleven additional property axes simultaneously.

\begin{figure}[ht]
    \centering
    \includegraphics[width=1.0\linewidth]{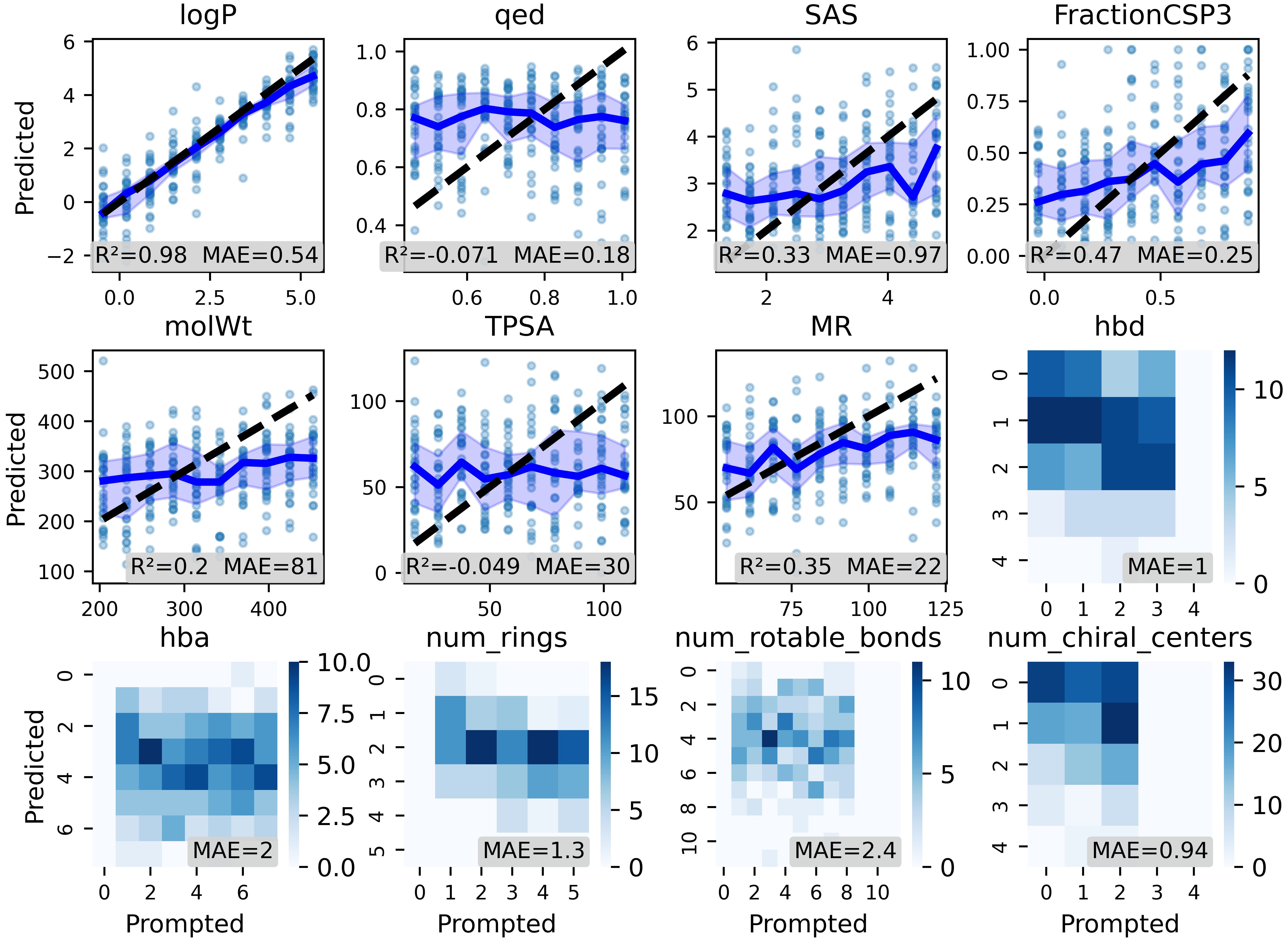}
    \caption{Calibration of conditional generation across twelve molecular properties for MolMiner-logP (1D specialist). Continuous properties show predicted versus prompted values with median and 25--75 quantile bands; discrete properties are summarized as confusion matrices. Mean Absolute Error (MAE) is reported for all properties; for continuous properties, the coefficient of determination $R^2$ is additionally reported between the median predicted and prompted values.}
    \label{fig:si-generation-calibration}
\end{figure}

Figure~\ref{fig:si-generation-calibration} reports the calibration of MolMiner-logP on the same property-by-property stress test used in Section~4.2 of the main text. On its single trained property, MolMiner-logP improves over the 12D generalist, reaching $R^2 = 0.98$ and MAE $= 0.54$ on logP, compared to $R^2 = 0.89$ and MAE $= 1.0$ for the 12D variant. This confirms the expected specialist advantage on the property it was trained to control. However, controllability collapses across the remaining eleven properties: SAS drops from $R^2 = 0.98$ to $0.33$, FractionCSP3 from $0.98$ to $0.47$, TPSA from $0.82$ to near zero, and QED, molWt, and MR all degrade substantially. The discrete properties (hbd, hba, num\_rings, num\_rotatable\_bonds, num\_chiral\_centers) similarly lose their diagonally-dominant confusion structure.

Taken together, the two evaluations characterize the generalist-vs-specialist trade-off from both directions. On the property it is trained to control, the specialist achieves a modest improvement in calibration (logP: $R^2$ $0.89 \rightarrow 0.98$) and somewhat tighter unconditional marginals across most properties. In exchange, it loses meaningful controllability over the eleven properties it was not trained on, with most calibration $R^2$ values falling below $0.5$ and several reaching values consistent with no control. The 12D generalist accepts a small distributional and per-property cost in order to provide simultaneous, calibrated control over twelve correlated property axes from partial user specifications, a capability the specialist does not support by construction.

\end{document}